\documentclass{article}

\usepackage[preprint]{./Styles/neurips_2025}

\usepackage[utf8]{inputenc} 
\usepackage[T1]{fontenc}    
\usepackage{hyperref}       
\usepackage{url}            
\usepackage{booktabs}       
\usepackage{amsfonts}       
\usepackage{nicefrac}       
\usepackage{microtype}      
\usepackage{amsmath}
\usepackage{bbm}
\usepackage{algorithm,algorithmicx,algpseudocode}

\usepackage[table,xcdraw]{xcolor}
\usepackage[normalem]{ulem}
\usepackage{float}
\useunder{\uline}{\ul}{}
\usepackage{wrapfig} 
\usepackage{graphicx}
\usepackage{lipsum} 
\usepackage{enumitem}

\newcommand{\method}{VARD w/o KL\xspace}
\newcommand{\methodKL}{VARD\xspace}

\definecolor{Highlight}{rgb}{0.89,0.89,0.94}

\usepackage{xspace}
\usepackage{amsmath}

\newcommand{\SE}{\mathrm{SE}}
\newcommand{\SO}{\mathrm{SO}}
\makeatletter
\DeclareRobustCommand\onedot{\futurelet\@let@token\@onedot}
\def\@onedot{\ifx\@let@token.\else.\null\fi\xspace}

\newcommand{\customfootnotetext}[2]{{%
  \renewcommand{\thefootnote}{#1}%
  \footnotetext[0]{#2}}}%

\def\eg{\textit{e.g}\onedot} 
\def\ie{\textit{i.e}\onedot}

\title{VARD: Efficient and Dense Fine-Tuning for Diffusion Models with Value-based RL}

\author{
  Fengyuan Dai*$^{1, 2}$ \\
  \And
  Zifeng Zhuang*$^{2}$ \\
  \And
  Yufei Huang$^{2}$ \\
  \And
  Siteng Huang$^{1}$ \\
  \And
  Bangyan Liao$^{2}$ \\
  \And
  Donglin Wang$^{2}$
  \quad
  Fajie Yuan$\dagger^{2}$
  \\
  \\
\noindent\hspace*{-0.025\linewidth}\parbox{1.00\linewidth}{
\centering
$^1$Zhejiang University, $^2$Westlake University
}
}

\begin{document}

\customfootnotetext{}{$^*$ Equal contribution; $^\dagger$  Corresponding author.
}

\maketitle

\begin{abstract}


Diffusion models have emerged as powerful generative tools across various domains, yet tailoring pre-trained models to exhibit specific desirable properties remains challenging. While reinforcement learning (RL) offers a promising solution, 
current methods struggle to simultaneously achieve stable, efficient fine-tuning and support non-differentiable rewards.
Furthermore, their reliance on sparse rewards provides inadequate supervision during intermediate steps, often resulting in suboptimal generation quality.
To address these limitations, dense and differentiable signals are required throughout the diffusion process. 
Hence, we propose \textbf{VA}lue-based \textbf{R}einforced \textbf{D}iffusion (VARD): a novel approach that first learns a value function predicting expection of rewards from intermediate states, and subsequently uses this value function with KL regularization to provide dense supervision throughout the generation process. Our method maintains proximity to the pretrained model while enabling effective and stable training via backpropagation.
Experimental results demonstrate that our approach facilitates better trajectory guidance, improves training efficiency and extends the applicability of RL to diffusion models optimized for complex, non-differentiable reward functions.


\end{abstract}
\section{Introduction}


Diffusion models~\citep{song2020denoising,song2020score,sohl2015deep} are generative models that synthesize data through iterative refinement of random noise via learned denoising steps. This framework has demonstrated broad applicability across multiple modalities, including image generation~\citep{rombach2022high}, text synthesis~\citep{gong2022diffuseq}, and protein design~\citep{ahern2025atom,wang2024diffusion}.
However, pre-trained diffusion models trained on large-scale datasets may lack specific desired properties. 
These limitations motivate optimizing target properties scored by reward models, a paradigm naturally suited to reinforcement learning (RL) frameworks.




While recent studies~\citep{miao2024training,uehara2024bridging,zhang2024large,lee2023aligning,lee2024parrot} have made significant advancements in integrating RL with diffusion model training, existing approaches confront a training strategy dilemma: non-differentiable reward vs. stable and efficient training.
Specifically, policy gradient approaches~\citep{black2023training, fan2023dpok} presuppose non-differentiable reward functions, inherently suffering from sample inefficiency and training instability due to high-variance gradient estimations~\citep{liu2025efficient, Deng_2024_CVPR, jia2025rewardfinetuningtwostepdiffusion}. 
Conversely, while reward backpropagation methods~\citep{clark2023directly} typically achieve more stable and efficient optimization, they are fundamentally constrained to differentiable reward scenarios, thus limiting their use in non-differentiable reward optimization tasks.

Furthermore, previous methods intuitively formulate these fine-tuning processes as RL problems with sparse rewards, \ie allocating meaningful rewards exclusively to the final denoising step while leaving intermediate states entirely unrewarded. 
This formulation overemphasizes the final output, resulting in inadequate supervision over the diffusion process itself. 
Lacking intermediate guidance, the model is compelled to generalize these coarse-grained reward signals across the entire diffusion process. Such generalization readily leads to error accumulation and hinders convergence speed, ultimately yielding suboptimal performance.




Inspired by the field of large language models, where process reward models (PRMs) predict intermediate rewards~\citep{lightman2023let,luo2024improve}, we explore a similar concept for diffusion models: a PRM designed to generate dense, differentiable feedback~\citep{zhang2024aligning, zhang2024confronting} across the entire diffusion trajectory, thereby overcoming the two aforementioned issues.
We observe that obtaining such a PRM may be particularly suitable for diffusion models because, unlike left-to-right natural language generation where semantics can shift significantly mid-sequence, diffusion trajectories generally maintain greater consistency throughout the process. This observation motivates us to train a PRM, analogous to a value function from an RL perspective, to predict the expectation of the final reward given the current state within the diffusion trajectory.


\begin{figure}[t]
\centering
\includegraphics[scale=0.32]{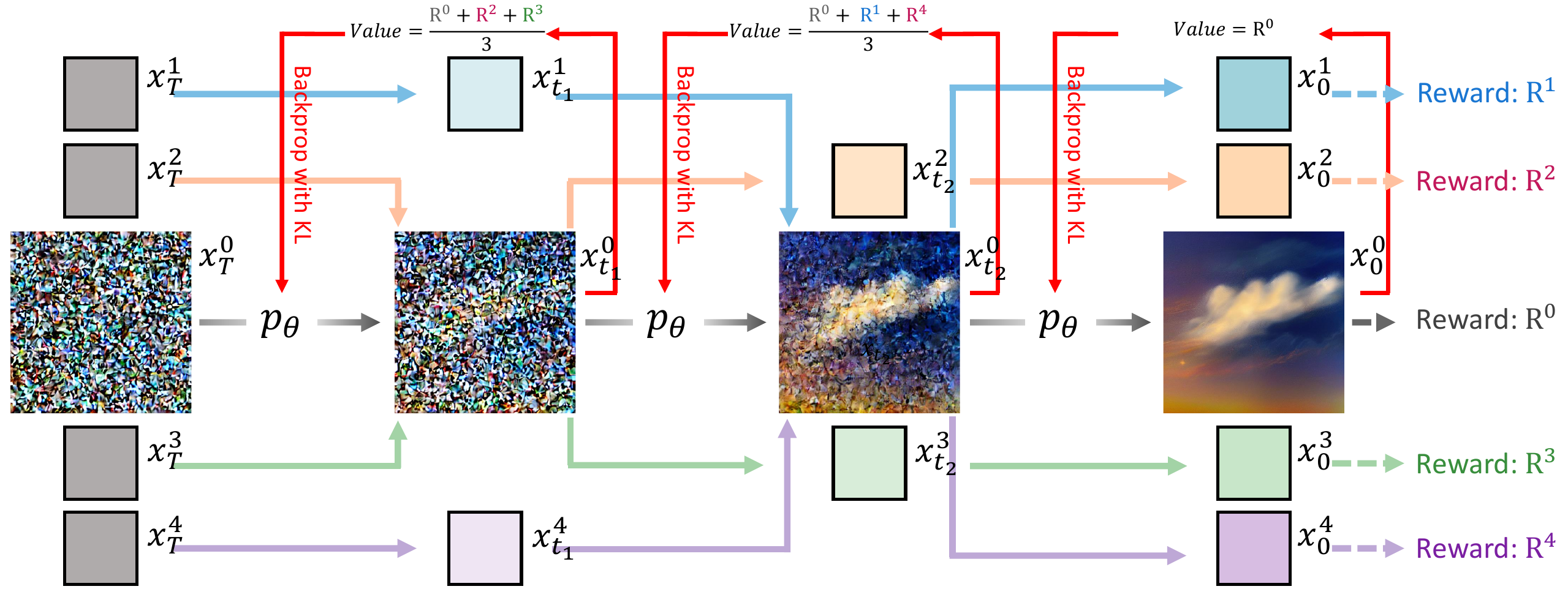}
\caption{\textbf{Illustration of the proposed VARD.} 
Different colors represent distinct diffusion trajectories, indexed by the superscripts on states ($x^i_t$) and final rewards ($R^i$). At intermediate steps, VARD's learned PRM calculates the value as the expected final reward, estimated by averaging $R_i$ of relevant trajectories considered to pass through the current state representation. 
For instance, as depicted for the intermediate state representation $x_{t_1}^1$, the value is computed as $\frac{R^3+R^2+R^0}{3}$, thereby aggregating information from trajectories 0, 2, and 3. 
This predicted value serves as a dense, global supervisory signal, utilized alongside KL regularization for  backpropagation throughout the diffusion process.
}
\label{figure:intro}
\end{figure}
Motivated by the analysis above, we propose our method, \textbf{VA}lue-based \textbf{R}einforced \textbf{D}iffusion (VARD), as shown in Figure~\ref{figure:intro}. 
For each intermediate state $x_t$, the learned PRM offers a global signal by predicting the average final reward expected from all complete diffusion trajectories that traverse through $x_t$.
Crucially, this prediction also serves as look-ahead information, guiding the diffusion process on how to proceed towards high-reward final states.
Using this dense supervision, VARD effectively and stably optimizes the diffusion model via backpropagation, complemented by a KL regularization term that prevents significant divergence from the pre-trained model's distribution.

Experimental validation of VARD on generative models for both protein structure and text-to-image synthesis reveals several key advantages.
Firstly, by providing dense supervision across the entire diffusion trajectory, VARD achieves faster convergence and improved sample efficiency. Furthermore, its trajectory-wise reward alignment, stabilized by KL regularization—in contrast to approaches that only consider final-state rewards—results in enhanced data quality. 
Finally, VARD also extends to non-differentiable rewards, as its integrated PRM decouples training from needing an externally differentiable reward function.
In summary, our contributions are as follows:
\begin{enumerate}
    \item We address the challenges of sparse reward supervision and the dilemma between stable optimization and non-differentiable reward functions by using a learned PRM. This PRM provides dense, differentiable reward signals throughout the entire diffusion process.
    \item We introduce VARD, a novel framework for fine-tuning diffusion models that integrates value as trajectory-wise supervision with KL regularization, thereby enhancing reward alignment while ensuring proximity to the pre-trained model.
    \item Extensive experiments demonstrate VARD's significant advantages: enhanced sample quality with improved efficiency, effective mitigation of reward hacking, and broad applicability, especially for scenarios with non-differentiable reward functions.
\end{enumerate}





\section{Preliminary}
\subsection{Diffusion Models}
We use the denoising diffusion probabilistic models (DDPMs)~\citep{ho2020denoising} and flow matching~\citep{lipman2022flow} for data generation. Here we take text-to-image generation for example; for specifics on protein backbone generation, please refer to Appendix~\ref{preliminary_protein}. Let  $q_{0}$  be the image data distribution $\mathbf{x}_{0} \sim q_{0}\left(\mathbf{x}_{0}\right), \mathbf{x}_{0} \in \mathbb{R}^{n}$. 
A DDPM aims to approximates  $q_{0}$  with a parameterized model  $p_{\theta}\left(\mathbf{x}_{0}\right)=\int \left[p_{T}\left(\mathbf{x}_{T}\right) \prod_{t=1}^{T} p_{\theta}\left(\mathbf{x}_{t-1} \mid \mathbf{x}_{t}\right) \right]d \mathbf{x}_{1: T}$.  The reverse process, also called denoising, is a Markov chain with the following initial distribution and dynamics:
\begin{align}\label{eq:reverse}
    p\left(\mathbf{x}_{T}\right)=\mathcal{N}(0, I), \quad p_{\theta}\left(\mathbf{x}_{t-1} \mid \mathbf{x}_{t}\right)=\mathcal{N}\left(\mu_{\theta}\left(\mathbf{x}_{t}, t\right), \Sigma_{t}\right).
\end{align}

The forward or diffusion process of DDPMs utilizes an approximate posterior  $q\left(\mathbf{x}_{1: T} \mid \mathbf{x}_{0}\right)$, which can be modeled as a Markov chain that incrementally adds Gaussian noise to the data according to a predifined variance schedule  $\beta_{1}, \ldots, \beta_{T}$:
\begin{align}
    q\left(\mathbf{x}_{1: T} \mid \mathbf{x}_{0}\right)=\prod_{t=1}^{T} q\left(\mathbf{x}_{t} \mid \mathbf{x}_{t-1}\right), \quad q\left(\mathbf{x}_{t} \mid \mathbf{x}_{t-1}\right)=\mathcal{N}\left(\sqrt{1-\beta_{t}} \mathbf{x}_{t-1}, \beta_{t} I\right).
\end{align}

Let  $\alpha_{t}=1-\beta_{t}, \bar{\alpha}_{t}=\prod_{s=1}^{t} \alpha_{s} $, and  $\tilde{\beta}_{t}=\frac{1-\bar{\alpha}_{t-1}}{1-\bar{\alpha}_{t}} \beta_{t} $. DDPMs adopt the parameterization  $\mu_{\theta}\left(\mathbf{x}_{t}, t\right)=\frac{1}{\sqrt{\alpha_{t}}}\left(\mathbf{x}_{t}-\frac{\beta_{t}}{\sqrt{1-\bar{\alpha}_{t}}} \epsilon_{\theta}\left(\mathbf{x}_{t}, t\right)\right) $ to represent the mean in Equation (\ref{eq:reverse}).

Training a DDPM is performed by optimizing a variational bound on the negative log-likelihood  $\mathbb{E}_{q}\left[-\log p_{\theta}\left(\mathbf{x}_{0}\right)\right] $, which is equivalent to optimizing
\begin{align}
   \mathbb{E}_{q}\left[\sum_{t=1}^{T} \operatorname{KL}\left(q\left(\mathbf{x}_{t-1} \mid \mathbf{x}_{t}, \mathbf{x}_{0}\right) \| p_{\theta}\left(\mathbf{x}_{t-1} \mid \mathbf{x}_{t}\right)\right)\right].
\end{align}
Note that the variance sequence $\left(\beta_{t}\right)_{t=1}^{T} \in(0,1)^{T}$ is chosen so that $\bar{\alpha}_{T} \approx 0$ is made to make $q\left(\mathbf{x}_{T} \mid \mathbf{x}_{0}\right) \approx   \mathcal{N}(0, I) $. The covariance matrix $\Sigma_{t}$ in Equation \ref{eq:reverse} is often set to  $\sigma_{t}^{2} I $, where  $\sigma_{t}^{2}$  is either  $\beta_{t}$  or  $\tilde{\beta}_{t}$, which is not trainable. 

Furthermore, diffusion models also excel at conditional data generation such as text-to-image generation with classifier-free guidance. Given context prompt $\mathbf{c} \sim P\left(\mathbf{c}\right)$, the image distribution becomes $q_{0}\left(\mathbf{x}_{0}, \mathbf{c}\right)$. During training, the forward noising process is used regardless of input $\mathbf{c}$. Both unconditional $\epsilon_{\theta}\left(\mathbf{x}_{t}, t\right)$ and conditional $\epsilon_{\theta}\left(\mathbf{x}_{t}, t, \mathbf{c}\right)$ are learned together. At test time, given a text prompt $\mathbf{c}$, the model generates conditional data according to $p_{\theta}\left(\mathbf{x}_{0}| \mathbf{c}\right)$.

\subsection{Markov Decision Processes and Offline Reinforcement Learning}
Reinforcement learning (RL)~\citep{sutton1998introduction} is a sequential decision making problem formulated by a Markov Decision Processes (MDP). MDP is usually defined by the tuple $\mathcal{M}=\left\{\mathcal{S}, \mathcal{A}, \rho_0, \mathcal{P}, \mathcal{R}\right\}$ in which $\mathcal{S}$ is the state space, $\mathcal{A}$ is the action space, $\rho_0$ is the initial distribution, $\mathcal{P}$ is the transition dynamics and $\mathcal{R}$ is the reward function. The policy $\pi$ interacts with the environment $\mathcal{M}$ to generate one trajectory $\tau = \left(\mathbf{s}_0, \mathbf{a}_0,\mathbf{s}_1, \mathbf{a}_1, \cdots, \mathbf{s}_T, \mathbf{a}_T\right)$ and its distribution is defined as $\mathcal{P}_{\pi}\left(\tau\right)$. Then the objective of RL is to derive one policy that can maximize the expectation under the trajectory distribution:
$\pi = \arg\max_{\pi} \mathbb{E}_{\tau \sim \mathcal{P}_{\pi}\left(\tau\right) }\left[\sum_{t=0}^T \mathcal{R}\left(\mathbf{s}_t, \mathbf{a}_t\right)\right].$

Offline RL~\citep{levine2020offline,fujimoto2021minimalist,zhuang2023behavior} forbids the interaction with the environment $\mathcal{M}$.  And only the offline dataset $\mathcal{D}=\left\{\mathbf{s}_t, \mathbf{a}_t, \mathbf{s}_{t+1}, r_t\right\}$ generated by the behavior policy $\pi_{\beta}$ is provided. Beyond return maximization, offline RL imposes an additional constraint that the learned policy must not deviate significantly from the behavior policy $\pi_{\beta}$:
\begin{align}
    \pi = \arg\max_{\pi} \mathbb{E}_{\mathbf{s}_t \sim \mathcal{D}, \mathbf{a}_t \sim \pi\left(\cdot|\mathbf{s}_t\right) }\left[\sum_{t=0}^T \mathcal{R}\left(\mathbf{s}_t, \mathbf{a}_t\right)\right] - D_{KL}\left(\pi\|\pi_{\beta}\right).
\end{align}

\section{Method}
In this section, we first formulate the problem of reinforcement learning fine-tuning of diffusion models and analyze its essence to reveal the inherent challenges during RL fine-tuning. Furthermore, we propose a minimalist approach for RL fine-tuning of diffusion called Value-based Reinforced Diffusion (VARD) and highlight the benefits of introduction of value function.
Finally, we present the details of the implementation and the overall algorithm.

\subsection{Reinforcement Learning Fine-tuning of Diffusion Models}
The generation process of conditional diffusion models, also the denoising process, can be modeled as a Markov Decision Process (MDP) with the following specifications:
$\mathcal{M}=\left\{\mathcal{S}, \mathcal{A}, \rho_0, \mathcal{P}, \mathcal{R}\right\}$
\begin{align}
    \nonumber \text{State space: }& \mathcal{S} =\left\{ \mathbf{s}_t|\mathbf{s}_t=\left[\mathbf{x}_{T-t}, \mathbf{c}\right]\right\}; \\
    \nonumber \text{Action space: }& \mathcal{A}=\left\{ \mathbf{a}_t|\mathbf{a}_t=\mathbf{x}_{T-t-1}\right\}; \\
    \nonumber \text{Initial distribution: }& \rho_0=\left[\mathcal{N}\left(0,I\right), p\left(\mathbf{c}\right)\right]; \\
    \nonumber \text{Transition dynamics: }& \mathcal{P}\left(\mathbf{s}_{t+1}|\mathbf{s}_t, \mathbf{a}_t\right) =\left[\delta_{\mathbf{c}},\delta_{\mathbf{x}_{T-t-1}}\right];\\
    \nonumber \text{Reward function: }& \mathcal{R}\left(\mathbf{s}_t, \mathbf{a}_t\right) = \mathcal{R}\left(\left[\mathbf{x}_{T-t},\mathbf{c}\right], \mathbf{x}_{T-t-1}\right).
\end{align}
Here $\delta_{\mathbf{c}},\delta_{\mathbf{x}_{T-t-1}}$ are the Dirac distributions with non-zero density only at $\mathbf{c}$ and $\mathbf{x}_{T-t-1}$.  Under this definition of MDP, the policy is $\pi_{\theta}\left(\mathbf{a}_t|\mathbf{s}_t\right)=p_{\theta}\left(\mathbf{x}_{T-t-1} | \mathbf{x}_{T-t}, \mathbf{c}\right)$. The goal of RL Fine-tuning is to maximize the expected cumulative rewards while close to the pretrained diffusion model:
\begin{align}\label{eq:RL-diffusion}
    J\left(\theta\right)=-\mathbb{E}_{p\left(\mathbf{c}\right)}\mathbb{E}_{p_{\theta}\left(\cdot|\mathbf{x}_{T-t}, \mathbf{c}\right)}\left[\sum_{t=0}^T\mathcal{R}\left(\left[\mathbf{x}_{T-t},\mathbf{c}\right], \mathbf{x}_{T-t-1}\right)\right] + D_{KL}\left(p_{\theta}\|p_{\theta_0}\right),
\end{align}
 where $p_{\theta_0}$ represents the pretrained diffusion model. To minimize this objective, two issues related to the reward function need to be considered:
\paragraph{1) Non-differentiable Rewards:} Sometimes, reward functions are non-differentiable, which precludes the direct optimization. One classical approach to address this issue is Policy Gradient (PG)~\citep{sutton1999policy}, which transforms return maximization into return-weighted maximum likelihood estimation. While this approach is general, it introduces relative complexity and cannot directly provide reward signals to the target diffusion model for optimization. In contrast, when the reward function is differentiable, return maximization can be achieved through direct optimization-analogous to Deterministic Policy Gradient (DPG)~\citep{silver2014deterministic} methods in RL. 

\paragraph{2) Sparse Rewards:} Strictly speaking, during the generation process, only the final step produces the actual samples, while all intermediate steps consist of a mixture of noise and the target samples. This means that the reward can only be obtained at the very last step:
\begin{align}
    \text{\texttt{Sparse} Reward function: }& \mathcal{R}\left(\mathbf{s}_t, \mathbf{a}_t\right) = \mathcal{R}\left(\left[\mathbf{x}_{T-t},\mathbf{c}\right], \mathbf{x}_{T-t-1}\right) = \mathbbm{1}\left(t=T-1\right) \cdot r\left(\mathbf{x}_0,\mathbf{c}\right).
\end{align}
From a reinforcement learning (RL) perspective, this presents a sparse reward problem, which is notoriously difficult to solve due to credit assignment difficulty~\citep{seo2019rewards} and insufficient learning signal~\citep{hare2019dealing}.
Additionally, drawing from the experience of RL fine-tuning for large language models (LLMs), the existing reward function resembles a classic Output Reward Model (ORM). A straightforward solution \textbf{to overcome the sparse reward challenge involves introducing a Process Reward Model (PRM)}\citep{luo2024improve}.



\subsection{Our Minimalist Approach}

To address this sparse reward challenge, a common approach in RL involves computing value functions to redistribute reward signals across intermediate steps that lack explicit feedback. This aligns with the Process Reward Model (PRM) paradigm in RL Fine-tuning for LLM, which emphasizes not just the final output but also the entire generation process. 
In RL Fine-tuning for diffusion models, we define this value function or PRM  as follows: 
\begin{align}\label{eq:value-def}
    V_{\phi}\left(\mathbf{x}_{T-t}, \mathbf{c}\right) := \mathbb{E}_{t \in \left[0,T\right]}\left[\sum_{t^{\prime}=t}^{T} \mathcal{R}\left(\left[\mathbf{x}_{T-t^{\prime}},\mathbf{c}\right], \mathbf{x}_{T-t^{\prime}-1}\right) | \mathbf{x}_{T-t^{\prime}}\right].
\end{align}
This value function represents the expected reward of images $\mathbf{x}_0$ generated by $p_{\theta}$ given the current sample $\mathbf{x}_{T-t}$.  For $\mathbf{x}_{T-t}$, this value function takes into account the rewards of all possible images that may be generated subsequently, such as Figure \ref{figure:intro}, effectively avoiding the interference of extreme rewards.
Guided by the value function, the diffusion model can directly navigate to high-value regions (also high reward) during the generation process, efficiently avoids unnecessary and potentially unproductive exploration.

An unbiased value function estimation method based on Monte Carlo approaches operates as follows:
\begin{align}\label{eq:value-loss}
   J\left(\phi\right) =  \mathbb{E}_{p\left(\mathbf{c}\right)}\mathbb{E}_{t \in \left[0,T\right]}\left[r\left(\mathbf{x}_{0}, \mathbf{c}\right)-V_{\phi}\left(\mathbf{x}_{T-t}, \mathbf{c}\right)\right]^2.
\end{align}
This value estimation technique leverages the neural network's inherent generalization capability to distribute the final step's reward signal across intermediate steps, effectively circumventing interference from artificial priors. This value function not only provides additional guidance for the generation process but is also differentiable, enabling direct optimization of $p_{\theta}$ using Deterministic Policy Gradient (DPG). Given the definition of value (\ref{eq:value-def}), the problem definition (\ref{eq:RL-diffusion}) can be formulated as the following VARD objective: 
\begin{align}\label{eq:VARD}
    J\left(\theta\right)=-\mathbb{E}_{p\left(\mathbf{c}\right)}\mathbb{E}_{\left(\mathbf{x}, \mathbf{x}^0\right)_{T-t} \sim \left(p_{\theta}, p_{\theta^0}\right)\left( \cdot |\mathbf{x}_{T-t+1}, \mathbf{c} \right)}\left[V_{\phi}\left(\mathbf{x}_{T-t}, \mathbf{c}\right)+\eta \cdot \left(\mathbf{x}_{T-t} - \mathbf{x}^0_{T-t}\right)^2\right] ,
\end{align}
where $\left(\mathbf{x}, \mathbf{x}^0\right)_{T-t} \sim \left(p_{\theta}, p_{\theta^0}\right)\left( \cdot |\mathbf{x}_{T-t+1}, \mathbf{c} \right)$ is an abbreviation of $\mathbf{x}_{T-t} \sim p_{\theta}\left( \cdot |\mathbf{x}_{T-t+1}, \mathbf{c} \right),$ $ \mathbf{x}^0_{T-t} \sim  p_{\theta^0}\left( \cdot |\mathbf{x}_{T-t+1}, \mathbf{c} \right)$.
Meanwhile, a more straightforward implementation of KL divergence has been adopted. 
The KL divergence term, weighted by $\eta$, serves to prevent model collapse during RL optimization, which can be achieved by minimizing the gap between the output of  the RL-optimized diffusion model $p_{\theta}$ and the output of pre-trained $p_{\theta^0}$ at each timestep. This can also be supported by:

\paragraph{Lemma 1} Minimizing the gap between $\mathbf{x}_{T-t}$ and $\mathbf{x}^0_{T-t}$ is equivalent to minimizing the KL.
\begin{align}
\nabla_\theta \mathbb{E}_{p\left(\mathbf{c}\right)}\mathbb{E}_{\left(\mathbf{x}, \mathbf{x}^0\right)_{T-t} \sim \left(p_{\theta}, p_{\theta^0}\right)\left( \cdot |\mathbf{x}_{T-t+1}, \mathbf{c} \right)}\left[\left(\mathbf{x}_{T-t} - \mathbf{x}^0_{T-t}\right)^2\right] =  2 \sigma^2 \cdot \nabla_\theta D_{KL}\left(p_{\theta}\|p_{\theta^0}\right).
\end{align}

\algrenewcommand\algorithmicindent{0.5em}%
\begin{figure}[t]
\vspace{-1.8em}
\begin{minipage}[H]{0.495\textwidth}
\begin{algorithm}[H]
  \caption{Value Pretraining} \label{alg:value_training}
  \small
  \begin{algorithmic}[1]
      \State {\bfseries Require:} Pretrained diffusion model $p_{\theta^0}$, randomly initialized value function $V_{\phi}$
      \For{$\mathbf{c}\sim p\left(\mathbf{c}\right)$}
        \For{$\left(\mathbf{x}^0_{T}, \mathbf{x}^0_{T-1}, \cdots, \mathbf{x}^0_{0}\right) \sim p_{\theta^0}$}
        \State Calculate the sparse reward $r\left(\mathbf{x}^0_{0}, \mathbf{c}\right)$
        \vspace{.004in}
        \State Update value function $V_{\phi}$ using Equation (\ref{eq:value-loss})
        \EndFor
      \EndFor
  \end{algorithmic}
\end{algorithm}
\end{minipage}
\hfill
\begin{minipage}[H]{0.495\textwidth}
\begin{algorithm}[H]
  \caption{Diffusion Fine-Tuning} \label{alg:diffusion_Fine-tuning}
  \small
  \begin{algorithmic}[1]
      \State {\bfseries Require:} Pretrained value function $V_{\phi^0}$, pretrained and finetuned diffusion model $p_{\theta^0}, p_{\theta}$
      \For{$\mathbf{c}\sim p\left(\mathbf{c}\right)$}
        \For{$t \in \left[0, T\right]$}
        \State $\left(\mathbf{x}, \mathbf{x}^0\right)_{T-t} \sim \left(p_{\theta}, p_{\theta^0}\right)\left( \cdot |\mathbf{x}_{T-t+1}, \mathbf{c} \right)$
        \State Update $p_{\theta}$ using Equation (\ref{eq:VARD}), and finetune $V_{\phi^0}$ 
        \EndFor
      \EndFor
  \end{algorithmic}
\end{algorithm}
\end{minipage}
\vspace{-1.2em}
\end{figure}



\

\subsection{Implementation Details}
The training process of VARD (Value-based Reinforced Diffusion) consists of two stages: pretraining of the value function Algorithm \ref{alg:value_training} and fine-tuning of the diffusion model Algorithm \ref{alg:diffusion_Fine-tuning}:
\setlength{\leftmargini}{2em}
\begin{itemize}
    \item Initially, the entire trajectory is sampled according to the pretrained diffusion model $p_{\theta^0}$, and the corresponding value function $V_{\phi^0}$ of $p_{\theta^0}$ is trained using Equation (\ref{eq:value-loss}). Once the value function has converged, the fine-tuning stage of the diffusion model commences.
    \item During this stage, the diffusion model generates a new sample $\mathbf{x}_{T-t}$ at each step, and the value function provides feedback signals to optimize the diffusion model following Equation (\ref{eq:VARD}). 
    Notably, after the diffusion model updates, the prior value function can no longer accurately predict values for samples from the updated model. Thus, following such an update, the value function should also be fine-tuned accordingly, with minimal additional computational overhead.
\end{itemize}

\section{Experiments}
In this section, we aim to comprehensively evaluate the effectiveness of VARD. First, we verify VARD's capability in handling non-differentiable rewards, specifically on protein structure design and image compressibility control (Section~\ref{sec:non_differential_rewards}). 
Then, we evaluate VARD's performance on differentiable reward designed for human preference alignment, including Aesthetic Score, PickScore, and ImageReward (Section~\ref{sec:human_preference_learning}). 
In Section~\ref{sec:learning_beyond_the_reward}, we delve into VARD's generalization ability, focusing specifically on its performance with unseen prompts, across different reward models, and in maintaining fidelity to the prior distribution. Additional experiments are presented in Appendix~\ref{sec:ablation_study}.

\subsection{Non-differentiable rewards}
\label{sec:non_differential_rewards}

\begin{figure}[t]
\centering
\includegraphics[scale=0.6]{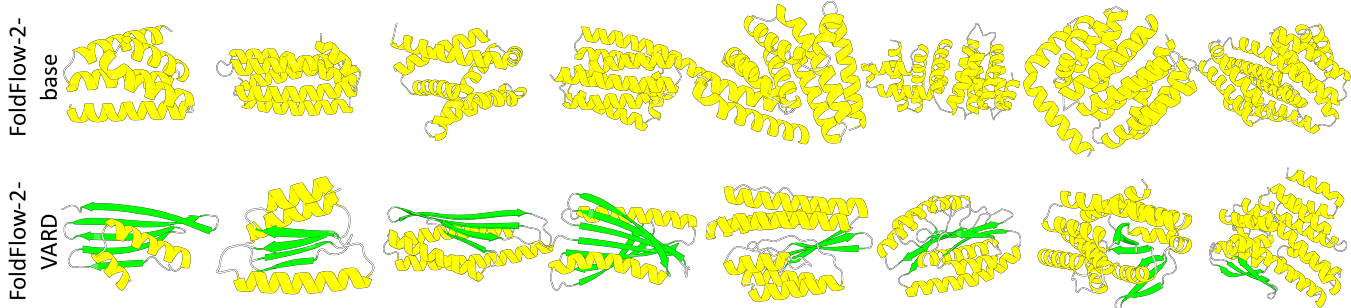}
\caption{\textbf{Qualitative analysis of generated protein structure from FoldFlow-2 base and VARD version.} Alpha-helices, beta-sheets and coils are colored \textcolor{yellow}{yellow}, \textcolor{green}{green} and \textcolor{gray}{gray}, respectively. The VARD version produces structures with significantly more beta-sheet regions, whereas the FoldFlow-2 base model generates almost exclusively alpha-helix regions.}
\label{figure:protein_visual}
\end{figure}

\textbf{Protein secondary structure diversity.} Current AI-generated protein models~\citep{watson2023novo} demonstrate systematic deviations in secondary structure distributions relative to native counterparts, with pronounced overrepresentation of helices~\citep{lin2024out, bose2023se, huguet2024sequence}.
\begin{wrapfigure}[15]{r}{0.53\textwidth} 
\vspace{-8pt}
\centering
\includegraphics[width=0.5\textwidth]{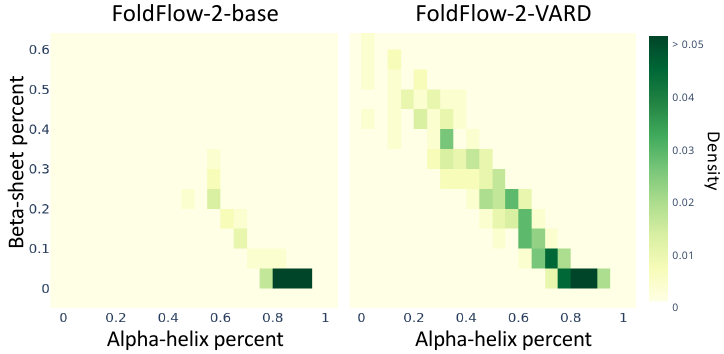} 
\vspace{-1.5mm}
\caption{\textbf{Secondary structure distribution comparison between FoldFlow-2-base and FoldFlow-2-VARD generated proteins.} Proteins are sampled at 50-residue increments between lengths 100-300, with 64 samples per length.}
\label{figure:protein_quan}
\vspace{-2mm}
\end{wrapfigure}
Motivated by this issue, we aim to fine-tune the pretrained SE(3) diffusion framework~\citep{yim2023se} to recalibrate conformational sampling. Following FoldFlow-2~\citep{huguet2024sequence}, we build a reward model defined as:
\begin{align}
    R=(\sum_{d\in D} p_dw_d) (1 + \sum_{d\in D} p_dlog p_d),
\end{align}
where $D=(a, b, c)$ represents secondary structure elements (alpha-helices, beta-sheets and coils). Here, we set $w_a=1,w_b =5,w_c=0.5$. We fine-tune the FoldFlow-2-base model using our proposed VARD approach, resulting in FoldFlow-2-VARD.

Figure~\ref{figure:protein_quan} reveals distinct differences in secondary structure distribution: FoldFlow-2-base demonstrates constrained beta-sheet formation, while our VARD-enhanced counterpart exhibits expanded structural diversity. This conclusion is qualitatively validated in Figure~\ref{figure:protein_visual}, where comparative analyses demonstrate the FoldFlow-2 model benefits from VARD fine-tuning and produces more beta-sheet.


\begin{figure}
\centering
\includegraphics[scale=0.39]{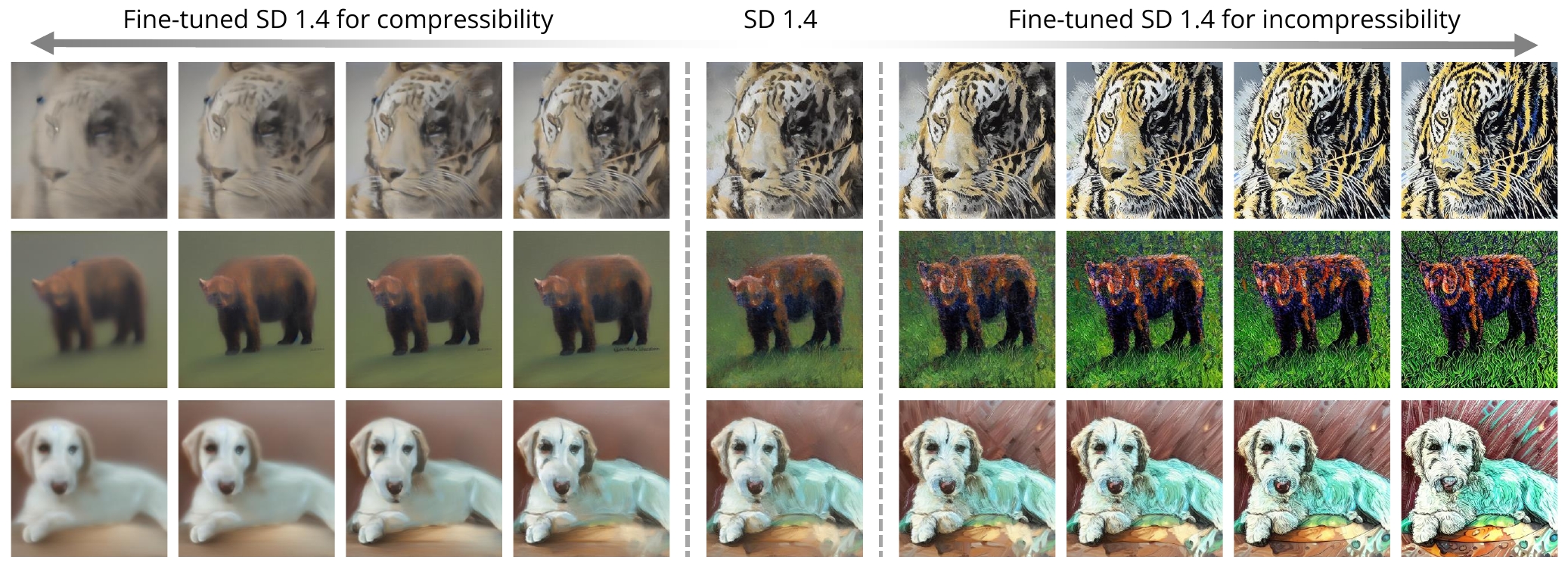}
\caption{
\textbf{Qualitative analysis of compressibility vs. incompressibility rewards.} Generated images correspond to checkpoints saved at 20-step intervals during training.
}
\label{figure:compress}

\end{figure}

\textbf{Compressibility and Incompressibility.} Following DDPO~\citep{black2023training}, we generate images with controlled JPEG compressibility. 
We formulate the reward scheme such that compressibility optimization minimizes the predicted size while incompressibility encourages larger file sizes.
The experiments are conducted on the 45 animal prompts. 
As shown in Figure~\ref{figure:compress}, when optimizing for compressibility, generated images exhibit progressive blurring and reduced diversity due to attenuated high-frequency regions. 
Conversely, maximizing compression difficulty yields images with enhanced textural complexity and photorealistic details, demonstrating VARD's competence in non-differentiable reward scenarios, where it achieves frequency-specific preference generation.

\subsection{Differentiable rewards}
\label{sec:human_preference_learning}

\begin{figure}[t]
\centering
\includegraphics[scale=0.95]{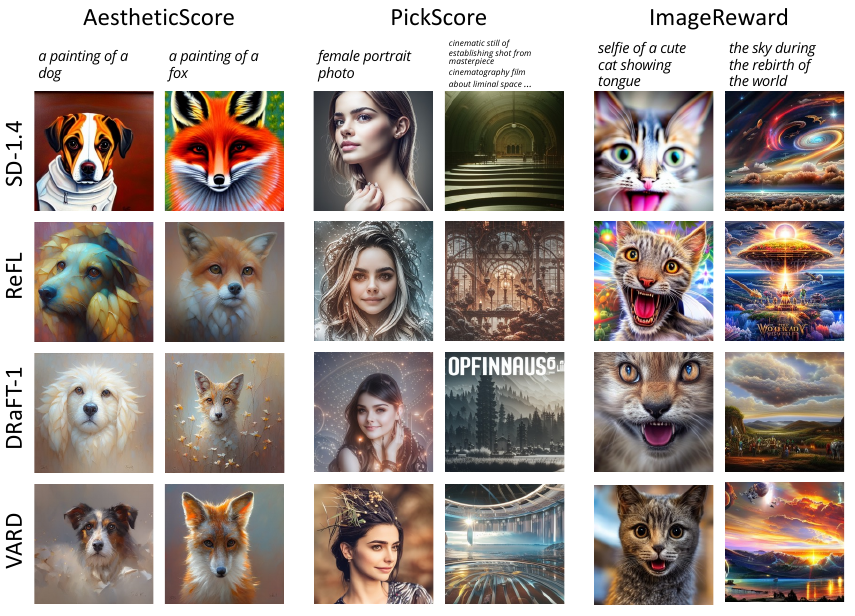}
\caption{
\textbf{Comparative visualization.} 
Compared to ReFL and DRaFT-1 aligned with specified reward preferences, \methodKL achieves demonstrably superior image quality and enhanced realism.
}
\label{figure:quantitive}
\end{figure}

\textbf{Experimental setup.}
We examine three reward functions grounded in human preference learning: Aesthetic Score~\citep{schuhmann2022laion}, PickScore~\citep{kirstain2023pick}, and ImageReward~\citep{xu2023imagereward}. Following DDPO~\citep{black2023training}, we evaluate Aesthetic Score using their standardized set of 45 animal prompts. For PickScore and ImageReward, we directly adopt their native training prompts (Pick-a-Pic and ImageRewardDB respectively) to preserve data distribution alignment. 

\textbf{Baselines.} 
We compare with two recent RL fine-tuning methods using direct reward backpropagation: ReFL~\citep{xu2023imagereward} and DRaFT-1~\citep{clark2023directly}. ReFL randomly samples the last 10 diffusion timesteps for gradients, while DRaFT applies stop-gradient to early timesteps and propagates through the final step. Notably, while DRaFT-1's original implementation integrates LoRA~\citep{hu2022lora} as an intrinsic component of their parameter-efficient framework, we intentionally adopt full model fine-tuning to isolate the effects of gradient propagation strategies.

\begin{figure}[t]
\vspace{-14pt}
\centering
\includegraphics[scale=0.65]{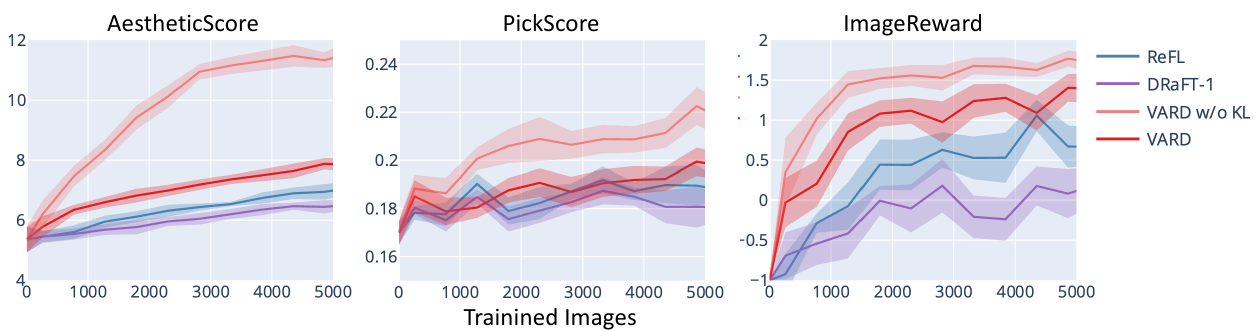}
\vspace{-2mm}
\caption{
\textbf{Reward v.s. sampled images.} Both \method and \methodKL attain higher reward values than baselines with substantially fewer training samples.
}
\label{figure:reward_trained_image}
\vspace{-12pt}
\end{figure}

\begin{wrapfigure}[23]{r}{0.45\textwidth}
    \centering
    \includegraphics[width=0.43\textwidth]{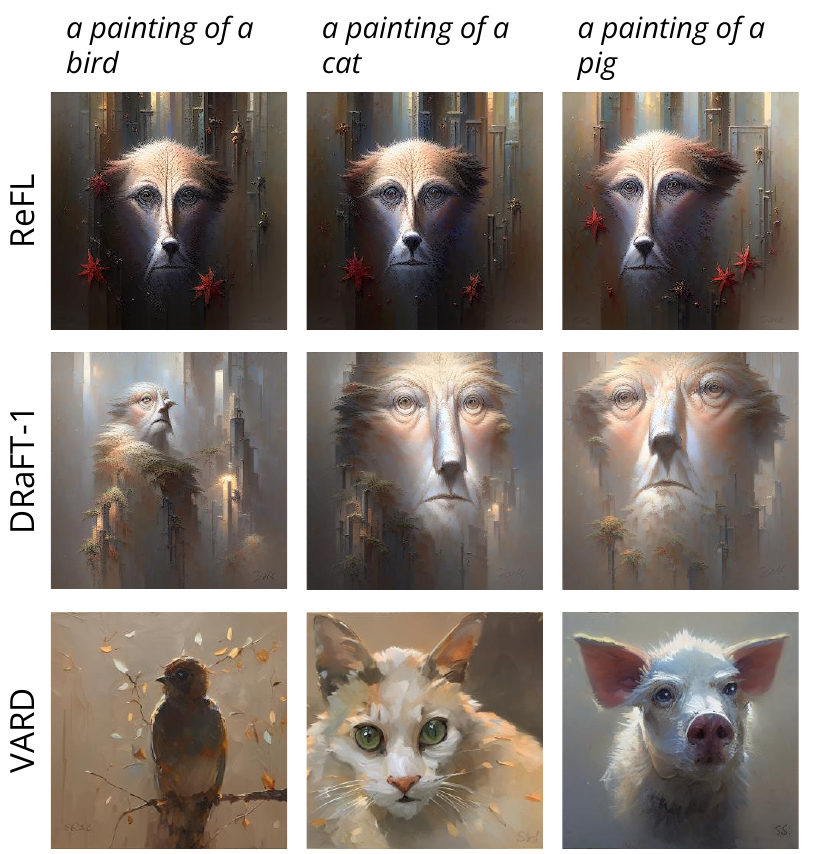} 
    \caption{\textbf{A example of reward hacking.} Directly backpropagation without KL catastrophically loses ability to generate prompt-aligned images, degenerating into texture-like noise patterns. All of the images is generated from checkpoints optimized with 500 steps.}
    \label{figure:reward_hacking}
\end{wrapfigure}

\textbf{Results.} 
Figure~\ref{figure:reward_trained_image} traces the relationship between training sample size and reward improvement, where \method and \methodKL exhibit steeper growth trajectories than baselines. The convergence curves confirm their superior sample efficiency across three  reward functions.
Complementary qualitative evidence is provided in Figure~\ref{figure:quantitive}. 
While all methods demonstrate improved sampled image quality, \methodKL achieves an optimal balance between reward maximization and distributional integrity through parameter-space anchoring near the pre-trained model. This strategic optimization results in significantly enhanced resistance to reward hacking compared to baselines and \method.

\subsection{Learning beyond the reward}
\label{sec:learning_beyond_the_reward}

As noted in \cite{gao2023scaling, liu2025efficient}, since reward functions serve as imperfect scorers, excessive focus on maximizing reward values through overoptimization can paradoxically degrade a model’s true performance. As shown in Figure~\ref{figure:reward_hacking}, previous reward backpropagation approaches frequently overlook this critical limitation, often leading to model collapse into a homogeneous and low-fidelity image that does not align with human expectations (\ie, reward hacking).

\begin{table}[b]
\vspace{-8pt}
\setlength{\tabcolsep}{5pt}
\centering
\caption{\textbf{Cross prompt generalization.} 
The highest scores are emphasized in boldface, with secondary best results indicated by single underlining.
}
\footnotesize
\begin{tabular}{@{}lccccc@{}}
\toprule
\textbf{Model}          & \textbf{Animation}  & \textbf{Concept Art} & \textbf{Painting}           & \textbf{Photo}      & \textbf{Averaged}   \\ \midrule
Stable Diffusion (SD 1.4)   & $0.2726$ &
                            $0.2661$ &
                            $0.2666$ &
                            $0.2727$ &
                            $0.2695$ \\
\midrule
DDPO                        & $0.2673$ & $0.2558$ & $0.2570$ & $0.2093$ & $0.2474$ \\
PRDP                        & $0.3223$ & $0.3175$ & $0.3172$ & $0.3159$ & $0.3182$ \\

ReFL                        & $0.3446$ & $0.3442$ & $0.3447$ & $0.3374$ & $0.3427$ \\
DRaFT-1                     & $0.3494$ & $0.3488$ & $0.3485$ & $0.3400$ & $0.3467$ \\
DRaFT-LV                    & $0.3551$ & $0.3552$ & $0.3560$ & $0.3480$ & $0.3536$ \\
\midrule
\rowcolor[rgb]{.949, .949, .949} 
\method                    & \underline{$0.3603$} & \underline{$0.3588$} & \underline{$0.3589$} & \underline{$0.3523$} & \underline{$0.3575$} \\
\rowcolor[rgb]{.949, .949, .949} 
\methodKL                 & \boldmath$0.3625$\unboldmath  &  \boldmath$0.3597$\unboldmath &  \boldmath$0.3597$\unboldmath &  \boldmath$0.3537$\unboldmath &  \boldmath$0.3589$\unboldmath \\

\bottomrule
\end{tabular}
\label{table:cross_prompts}
\end{table}


We investigate whether our method can effectively learn beyond the reward, ensuring that the model not only maximizes the reward function, but also captures generalizable and meaningful patterns that align with human expectations. Specifically, we employ three key evaluation settings: \textbf{unseen-prompt generalization}, \textbf{cross-reward validation}, and \textbf{ prior consistency-reward balance}.

\textbf{Unseen-prompt generalization.} 
Following the DRaFT~\citep{clark2023directly}, we first fine-tuned the model using the HPDv2~\citep{wu2023humanpreferencescorev2} training set with HPSv2 serving as the reward model. 
Subsequently, we evaluated the fine-tuned model's performance across all four test sets from HPDv2, using the HPSv2 score as the evaluation metric.

The evaluation results are presented in Table~\ref{table:cross_prompts}. Performance metrics for the baseline models are sourced from the DRaFT and PRDP~\citep{Deng_2024_CVPR} paper. 
DDPO fails to effectively optimize the reward function, whereas PRDP achieves only moderate reward values. Both methods underperform standard reward backpropagation approaches.
\method demonstrates superior unseen-prompt generalization by outperforming ReFL and DRaFT across all four test sets. Furthermore, \methodKL achieves the best performance, indicating that incorporation of KL regularization enhances reward learning for unseen prompts.

\begin{figure}[t]
\centering
\vspace{-12pt}
\includegraphics[scale=0.51]{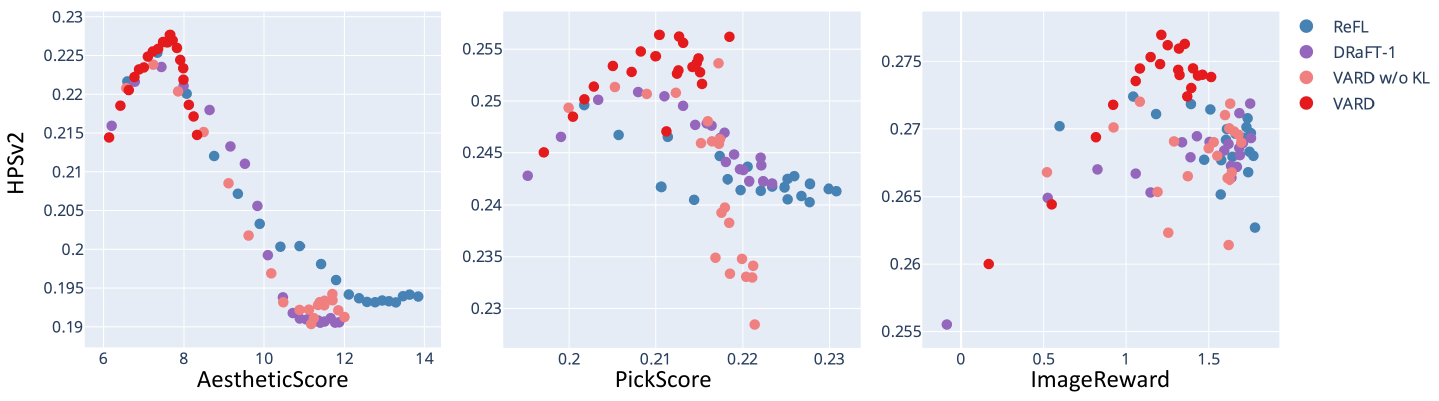}
\vspace{-2mm}
\caption{
\textbf{Evaluated by HPSv2.}  \methodKL consistently attains the highest HPSv2 scores while simultaneously preserving a modest reward.
}
\vspace{-4mm}
\label{figure:cross_reward}
\end{figure}

\begin{figure}[t]
\centering
\includegraphics[scale=0.51]{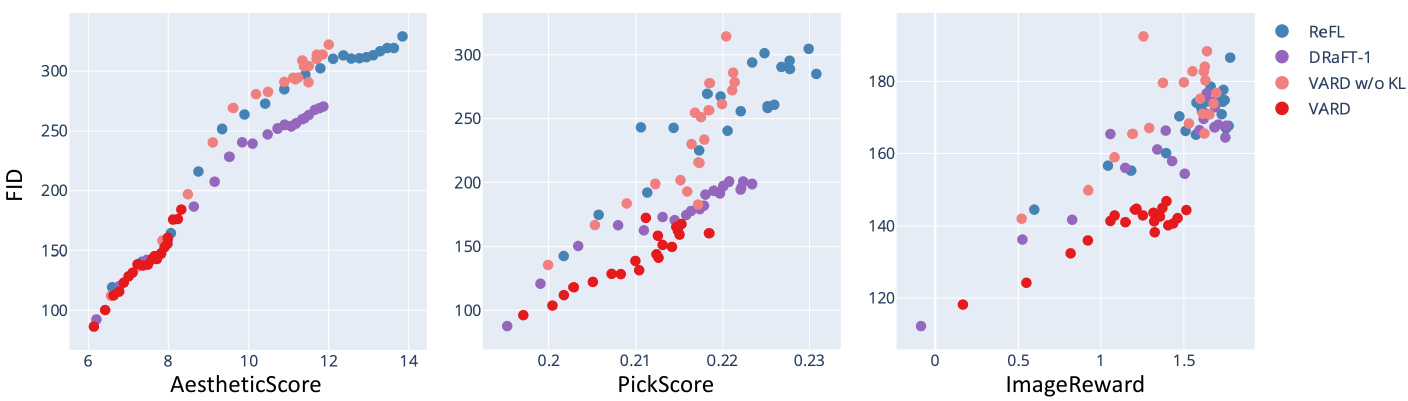}
\caption{
\textbf{Prior maintenance.} The image generated by \methodKL remains the closest to the base model as the reward increases.
}
\vspace{-4mm}
\label{figure:fid_reward}
\end{figure}

\textbf{Cross-reward validation.}
Next, we assess the performance of the learned model using other reward models as benchmarks. We used HPSv2 as the evaluation metric, as the Aesthetic Score, ImageReward, and PickScore are all grounded in human preference learning. We save model checkpoints every 50 steps and generate images for evaluation. 

The results of this comprehensive evaluation are illustrated in Figure~\ref{figure:cross_reward}. 
Methods lacking KL regularization are prone to overfitting to high-reward signals, often resulting in significantly lower HPSv2 scores. This phenomenon is particularly pronounced with Aesthetic Score, as it does not account for textual alignment and the datasets we trained on, \ie the animals, are relatively small, making it susceptible to model forgetting. In contrast, \methodKL consistently achieves the highest HPSv2 scores while maintaining a balanced reward level, demonstrating superior robustness and generalization compared to previous approaches. This highlights the effectiveness of our approach in mitigating overfitting and preserving alignment with human preferences.

\textbf{Prior consistency-reward balance.} 
We probe into the prior-preserving capabilities of the fine-tuned model. Specifically, we calculate the per-prompt FID score by comparing images generated from the pre-trained model with those from the fine-tuned model, and then average the FID scores across all prompts.
A higher FID score indicates a significant deviation from the pre-trained model.
As illustrated in Figure~\ref{figure:fid_reward}, models without KL tend to produce low-fidelity images as the reward increases. 
In contrast, \methodKL achieves substantial reward enhancement while maintaining a low FID score, thereby effectively balancing reward optimization with prior preservation.

\section{Related Work}

\textbf{RLHF and reward modeling in LLMs}. Reinforcement learning from human feedback (RLHF)~\citep{hou2024does, nika2024reward} has become a cornerstone in aligning LLMs with human preferences and improved reasoning, the reward model being a key component. 
For fine-grained supervision, especially in complex reasoning where the generation process matters, process reward models (PRMs) evaluate intermediate steps to guide the model's internal pathways~\citep{lightman2023let, luo2024improve, wang2023math}. This contrasts with outcome reward models (ORMs)~\citep{lyu2025exploring}, which assess the final generated output. 
Due to their relative simplicity in training, ORMs are widely adopted and become a mainstream technique for improving reasoning ability~\citep{guo2025deepseek}.

\textbf{RL funetuning for diffusion models}.
Training diffusion models with RL is an increasingly prominent approach 
in image generation~\citep{domingo2024adjoint} and AI for science field~\citep{uehara2025reward, uehara2024fine, uehara2024feedback, li2024derivative, uehara2024bridging}. 
Some methods utilize differentiable rewards, allowing for direct gradient backpropagation through parts of the diffusion sampling process for end-to-end optimization~\citep{xing2025focus, zhao2024adding, eyring2024reno}. 
Another thrust adapts from RLHF in LLMs, such as RL algorithms like PPO~\citep{schulman2017proximal} or DPO~\citep{rafailov2023direct}, often to effectively incorporate human feedback or navigate non-differentiable objectives~\citep{lee2024parrot, zhang2023towards, xu2024visionreward, wallace2024diffusion}.




\section{Conclusion}
In this paper, we introduce \textbf{VA}lue-based \textbf{R}einforced \textbf{D}iffusion (VARD), a novel approach for efficiently optimizing diffusion models with reward preference. By training a value function (also PRM) to provide dense supervision signals throughout the diffusion trajectory, VARD effectively overcomes the limitations of sparse rewards and the dilemma between differentiable rewards and stable training.
Our approach is validated on non-differentiable and differentiable rewards for human preference learning, as well as by exploring its generalization beyond simple reward maximization.

\section{Limitation}
The primary limitation of VARD stems from its reliance on the learned value function, whose accuracy critically influences the entire fine-tuning process.
This presents practical challenges. First, achieving sufficient accuracy for the value function sometimes necessitate a pre-training phase before it can effectively guide the diffusion model's training.
Second, concurrently training the network of the value function alongside the diffusion model, while enabling online adaptation of the value estimates, inevitably introduces additional GPU memory requirements. Nevertheless, we suggest these issues can be substantially mitigated by selecting appropriate value function architectures, which we will elaborate on later.
Looking ahead, future work will involve investigating and validating VARD's effectiveness when applied to more advanced and large-scale diffusion model architectures.

\section{Social impact}
Our research introduces VARD, a novel framework for aligning diffusion models with specific objectives using RL. The current study focuses primarily on algorithmic development and evaluation within controlled experimental settings, and does not involve direct real-world deployment or application studies. As a result, there are many potential societal consequences of our work, none of which we feel must be specifically highlighted here.

\newpage
\bibliography{reference}

\begin{thebibliography}{69}
\providecommand{\natexlab}[1]{#1}
\providecommand{\url}[1]{\texttt{#1}}
\expandafter\ifx\csname urlstyle\endcsname\relax
  \providecommand{\doi}[1]{doi: #1}\else
  \providecommand{\doi}{doi: \begingroup \urlstyle{rm}\Url}\fi

\bibitem[Ahern et~al.(2025)Ahern, Yim, Tischer, Salike, Woodbury, Kim, Kalvet,
  Kipnis, Coventry, Altae-Tran, et~al.]{ahern2025atom}
Woody Ahern, Jason Yim, Doug Tischer, Saman Salike, Seth Woodbury, Donghyo Kim,
  Indrek Kalvet, Yakov Kipnis, Brian Coventry, Han Altae-Tran, et~al.
\newblock Atom level enzyme active site scaffolding using rfdiffusion2.
\newblock \emph{bioRxiv}, pages 2025--04, 2025.

\bibitem[Black et~al.(2023)Black, Janner, Du, Kostrikov, and
  Levine]{black2023training}
Kevin Black, Michael Janner, Yilun Du, Ilya Kostrikov, and Sergey Levine.
\newblock Training diffusion models with reinforcement learning.
\newblock \emph{arXiv preprint arXiv:2305.13301}, 2023.

\bibitem[Bose et~al.(2023)Bose, Akhound-Sadegh, Huguet, Fatras, Rector-Brooks,
  Liu, Nica, Korablyov, Bronstein, and Tong]{bose2023se}
Avishek~Joey Bose, Tara Akhound-Sadegh, Guillaume Huguet, Kilian Fatras, Jarrid
  Rector-Brooks, Cheng-Hao Liu, Andrei~Cristian Nica, Maksym Korablyov, Michael
  Bronstein, and Alexander Tong.
\newblock Se (3)-stochastic flow matching for protein backbone generation.
\newblock \emph{arXiv preprint arXiv:2310.02391}, 2023.

\bibitem[Clark et~al.(2023)Clark, Vicol, Swersky, and Fleet]{clark2023directly}
Kevin Clark, Paul Vicol, Kevin Swersky, and David~J Fleet.
\newblock Directly fine-tuning diffusion models on differentiable rewards.
\newblock \emph{ICLR}, 2023.

\bibitem[Deng et~al.(2024)Deng, Wang, Wei, Hou, and Grundmann]{Deng_2024_CVPR}
Fei Deng, Qifei Wang, Wei Wei, Tingbo Hou, and Matthias Grundmann.
\newblock Prdp: Proximal reward difference prediction for large-scale reward
  finetuning of diffusion models.
\newblock In \emph{Proceedings of the IEEE/CVF Conference on Computer Vision
  and Pattern Recognition (CVPR)}, pages 7423--7433, June 2024.

\bibitem[Domingo-Enrich et~al.(2024)Domingo-Enrich, Drozdzal, Karrer, and
  Chen]{domingo2024adjoint}
Carles Domingo-Enrich, Michal Drozdzal, Brian Karrer, and Ricky~TQ Chen.
\newblock Adjoint matching: Fine-tuning flow and diffusion generative models
  with memoryless stochastic optimal control.
\newblock \emph{arXiv preprint arXiv:2409.08861}, 2024.

\bibitem[Eyring et~al.(2024)Eyring, Karthik, Roth, Dosovitskiy, and
  Akata]{eyring2024reno}
Luca Eyring, Shyamgopal Karthik, Karsten Roth, Alexey Dosovitskiy, and Zeynep
  Akata.
\newblock Reno: Enhancing one-step text-to-image models through reward-based
  noise optimization.
\newblock \emph{Advances in Neural Information Processing Systems},
  37:\penalty0 125487--125519, 2024.

\bibitem[Fan et~al.(2023)Fan, Watkins, Du, Liu, Ryu, Boutilier, Abbeel,
  Ghavamzadeh, Lee, and Lee]{fan2023dpok}
Ying Fan, Olivia Watkins, Yuqing Du, Hao Liu, Moonkyung Ryu, Craig Boutilier,
  Pieter Abbeel, Mohammad Ghavamzadeh, Kangwook Lee, and Kimin Lee.
\newblock Dpok: Reinforcement learning for fine-tuning text-to-image diffusion
  models.
\newblock \emph{Advances in Neural Information Processing Systems},
  36:\penalty0 79858--79885, 2023.

\bibitem[Fox et~al.(2014)Fox, Brenner, and Chandonia]{fox2014scope}
Naomi~K Fox, Steven~E Brenner, and John-Marc Chandonia.
\newblock Scope: Structural classification of proteins—extended, integrating
  scop and astral data and classification of new structures.
\newblock \emph{Nucleic acids research}, 42\penalty0 (D1):\penalty0 D304--D309,
  2014.

\bibitem[Fujimoto and Gu(2021)]{fujimoto2021minimalist}
Scott Fujimoto and Shixiang~Shane Gu.
\newblock A minimalist approach to offline reinforcement learning.
\newblock \emph{Advances in neural information processing systems},
  34:\penalty0 20132--20145, 2021.

\bibitem[Gao et~al.(2023)Gao, Schulman, and Hilton]{gao2023scaling}
Leo Gao, John Schulman, and Jacob Hilton.
\newblock Scaling laws for reward model overoptimization.
\newblock In \emph{International Conference on Machine Learning}, pages
  10835--10866. PMLR, 2023.

\bibitem[Gong et~al.(2022)Gong, Li, Feng, Wu, and Kong]{gong2022diffuseq}
Shansan Gong, Mukai Li, Jiangtao Feng, Zhiyong Wu, and LingPeng Kong.
\newblock Diffuseq: Sequence to sequence text generation with diffusion models.
\newblock \emph{arXiv preprint arXiv:2210.08933}, 2022.

\bibitem[Guo et~al.(2025)Guo, Yang, Zhang, Song, Zhang, Xu, Zhu, Ma, Wang, Bi,
  et~al.]{guo2025deepseek}
Daya Guo, Dejian Yang, Haowei Zhang, Junxiao Song, Ruoyu Zhang, Runxin Xu,
  Qihao Zhu, Shirong Ma, Peiyi Wang, Xiao Bi, et~al.
\newblock Deepseek-r1: Incentivizing reasoning capability in llms via
  reinforcement learning.
\newblock \emph{arXiv preprint arXiv:2501.12948}, 2025.

\bibitem[Hare(2019)]{hare2019dealing}
Joshua Hare.
\newblock Dealing with sparse rewards in reinforcement learning.
\newblock \emph{arXiv preprint arXiv:1910.09281}, 2019.

\bibitem[He et~al.(2016)He, Zhang, Ren, and Sun]{he2016deep}
Kaiming He, Xiangyu Zhang, Shaoqing Ren, and Jian Sun.
\newblock Deep residual learning for image recognition.
\newblock In \emph{Proceedings of the IEEE conference on computer vision and
  pattern recognition}, pages 770--778, 2016.

\bibitem[Ho et~al.(2020)Ho, Jain, and Abbeel]{ho2020denoising}
Jonathan Ho, Ajay Jain, and Pieter Abbeel.
\newblock Denoising diffusion probabilistic models.
\newblock \emph{Advances in neural information processing systems},
  33:\penalty0 6840--6851, 2020.

\bibitem[Hou et~al.(2024)Hou, Du, Niu, Du, Zeng, Liu, Huang, Wang, Tang, and
  Dong]{hou2024does}
Zhenyu Hou, Pengfan Du, Yilin Niu, Zhengxiao Du, Aohan Zeng, Xiao Liu, Minlie
  Huang, Hongning Wang, Jie Tang, and Yuxiao Dong.
\newblock Does rlhf scale? exploring the impacts from data, model, and method.
\newblock \emph{arXiv preprint arXiv:2412.06000}, 2024.

\bibitem[Hsu et~al.(2022)Hsu, Verkuil, Liu, Lin, Hie, Sercu, Lerer, and
  Rives]{hsu2022learning}
Chloe Hsu, Robert Verkuil, Jason Liu, Zeming Lin, Brian Hie, Tom Sercu, Adam
  Lerer, and Alexander Rives.
\newblock Learning inverse folding from millions of predicted structures.
\newblock In \emph{International conference on machine learning}, pages
  8946--8970. PMLR, 2022.

\bibitem[Hu et~al.(2022)Hu, Shen, Wallis, Allen-Zhu, Li, Wang, Wang, Chen,
  et~al.]{hu2022lora}
Edward~J Hu, Yelong Shen, Phillip Wallis, Zeyuan Allen-Zhu, Yuanzhi Li, Shean
  Wang, Lu~Wang, Weizhu Chen, et~al.
\newblock Lora: Low-rank adaptation of large language models.
\newblock \emph{ICLR}, 1\penalty0 (2):\penalty0 3, 2022.

\bibitem[Huguet et~al.(2024)Huguet, Vuckovic, Fatras, Thibodeau-Laufer, Lemos,
  Islam, Liu, Rector-Brooks, Akhound-Sadegh, Bronstein,
  et~al.]{huguet2024sequence}
Guillaume Huguet, James Vuckovic, Kilian Fatras, Eric Thibodeau-Laufer, Pablo
  Lemos, Riashat Islam, Cheng-Hao Liu, Jarrid Rector-Brooks, Tara
  Akhound-Sadegh, Michael Bronstein, et~al.
\newblock Sequence-augmented se (3)-flow matching for conditional protein
  backbone generation.
\newblock \emph{arXiv preprint arXiv:2405.20313}, 2024.

\bibitem[Jia et~al.(2025)Jia, Nan, Zhao, and
  Liu]{jia2025rewardfinetuningtwostepdiffusion}
Zhiwei Jia, Yuesong Nan, Huixi Zhao, and Gengdai Liu.
\newblock Reward fine-tuning two-step diffusion models via learning
  differentiable latent-space surrogate reward, 2025.
\newblock URL \url{https://arxiv.org/abs/2411.15247}.

\bibitem[Jumper et~al.(2021)Jumper, Evans, Pritzel, Green, Figurnov,
  Ronneberger, Tunyasuvunakool, Bates, {\v{Z}}{\'\i}dek, Potapenko,
  et~al.]{jumper2021highly}
John Jumper, Richard Evans, Alexander Pritzel, Tim Green, Michael Figurnov,
  Olaf Ronneberger, Kathryn Tunyasuvunakool, Russ Bates, Augustin
  {\v{Z}}{\'\i}dek, Anna Potapenko, et~al.
\newblock Highly accurate protein structure prediction with alphafold.
\newblock \emph{nature}, 596\penalty0 (7873):\penalty0 583--589, 2021.

\bibitem[Kirstain et~al.(2023)Kirstain, Polyak, Singer, Matiana, Penna, and
  Levy]{kirstain2023pick}
Yuval Kirstain, Adam Polyak, Uriel Singer, Shahbuland Matiana, Joe Penna, and
  Omer Levy.
\newblock Pick-a-pic: An open dataset of user preferences for text-to-image
  generation.
\newblock \emph{Advances in Neural Information Processing Systems},
  36:\penalty0 36652--36663, 2023.

\bibitem[Lee et~al.(2023)Lee, Liu, Ryu, Watkins, Du, Boutilier, Abbeel,
  Ghavamzadeh, and Gu]{lee2023aligning}
Kimin Lee, Hao Liu, Moonkyung Ryu, Olivia Watkins, Yuqing Du, Craig Boutilier,
  Pieter Abbeel, Mohammad Ghavamzadeh, and Shixiang~Shane Gu.
\newblock Aligning text-to-image models using human feedback.
\newblock \emph{arXiv preprint arXiv:2302.12192}, 2023.

\bibitem[Lee et~al.(2024)Lee, Li, Ke, Yoo, Zhang, Yu, Wang, Deng, Entis, He,
  et~al.]{lee2024parrot}
Seung~Hyun Lee, Yinxiao Li, Junjie Ke, Innfarn Yoo, Han Zhang, Jiahui Yu, Qifei
  Wang, Fei Deng, Glenn Entis, Junfeng He, et~al.
\newblock Parrot: Pareto-optimal multi-reward reinforcement learning framework
  for text-to-image generation.
\newblock In \emph{European Conference on Computer Vision}, pages 462--478.
  Springer, 2024.

\bibitem[Levine et~al.(2020)Levine, Kumar, Tucker, and Fu]{levine2020offline}
Sergey Levine, Aviral Kumar, George Tucker, and Justin Fu.
\newblock Offline reinforcement learning: Tutorial, review, and perspectives on
  open problems.
\newblock \emph{arXiv preprint arXiv:2005.01643}, 2020.

\bibitem[Li et~al.(2024)Li, Zhao, Wang, Scalia, Eraslan, Nair, Biancalani, Ji,
  Regev, Levine, et~al.]{li2024derivative}
Xiner Li, Yulai Zhao, Chenyu Wang, Gabriele Scalia, Gokcen Eraslan, Surag Nair,
  Tommaso Biancalani, Shuiwang Ji, Aviv Regev, Sergey Levine, et~al.
\newblock Derivative-free guidance in continuous and discrete diffusion models
  with soft value-based decoding.
\newblock \emph{arXiv preprint arXiv:2408.08252}, 2024.

\bibitem[Lightman et~al.(2023)Lightman, Kosaraju, Burda, Edwards, Baker, Lee,
  Leike, Schulman, Sutskever, and Cobbe]{lightman2023let}
Hunter Lightman, Vineet Kosaraju, Yuri Burda, Harrison Edwards, Bowen Baker,
  Teddy Lee, Jan Leike, John Schulman, Ilya Sutskever, and Karl Cobbe.
\newblock Let's verify step by step.
\newblock In \emph{The Twelfth International Conference on Learning
  Representations}, 2023.

\bibitem[Lin et~al.(2024)Lin, Lee, Zhang, and AlQuraishi]{lin2024out}
Yeqing Lin, Minji Lee, Zhao Zhang, and Mohammed AlQuraishi.
\newblock Out of many, one: Designing and scaffolding proteins at the scale of
  the structural universe with genie 2.
\newblock \emph{arXiv preprint arXiv:2405.15489}, 2024.

\bibitem[Lipman et~al.(2022)Lipman, Chen, Ben-Hamu, Nickel, and
  Le]{lipman2022flow}
Yaron Lipman, Ricky~TQ Chen, Heli Ben-Hamu, Maximilian Nickel, and Matt Le.
\newblock Flow matching for generative modeling.
\newblock \emph{arXiv preprint arXiv:2210.02747}, 2022.

\bibitem[Liu et~al.(2022)Liu, Gong, and Liu]{liu2022flow}
Xingchao Liu, Chengyue Gong, and Qiang Liu.
\newblock Flow straight and fast: Learning to generate and transfer data with
  rectified flow.
\newblock \emph{arXiv preprint arXiv:2209.03003}, 2022.

\bibitem[Liu et~al.(2025)Liu, Xiao, Liu, Bengio, and Zhang]{liu2025efficient}
Zhen Liu, Tim~Z. Xiao, Weiyang Liu, Yoshua Bengio, and Dinghuai Zhang.
\newblock Efficient diversity-preserving diffusion alignment via
  gradient-informed {GF}lownets.
\newblock In \emph{The Thirteenth International Conference on Learning
  Representations}, 2025.
\newblock URL \url{https://openreview.net/forum?id=Aye5wL6TCn}.

\bibitem[Loshchilov and Hutter(2017)]{loshchilov2017decoupled}
Ilya Loshchilov and Frank Hutter.
\newblock Decoupled weight decay regularization.
\newblock \emph{arXiv preprint arXiv:1711.05101}, 2017.

\bibitem[Luo et~al.(2024)Luo, Liu, Liu, Phatale, Guo, Lara, Li, Shu, Zhu, Meng,
  et~al.]{luo2024improve}
Liangchen Luo, Yinxiao Liu, Rosanne Liu, Samrat Phatale, Meiqi Guo, Harsh Lara,
  Yunxuan Li, Lei Shu, Yun Zhu, Lei Meng, et~al.
\newblock Improve mathematical reasoning in language models by automated
  process supervision.
\newblock \emph{arXiv preprint arXiv:2406.06592}, 2024.

\bibitem[Lyu et~al.(2025)Lyu, Gao, Gu, Zhang, Gao, Liu, Wang, Li, Zhao, Huang,
  et~al.]{lyu2025exploring}
Chengqi Lyu, Songyang Gao, Yuzhe Gu, Wenwei Zhang, Jianfei Gao, Kuikun Liu,
  Ziyi Wang, Shuaibin Li, Qian Zhao, Haian Huang, et~al.
\newblock Exploring the limit of outcome reward for learning mathematical
  reasoning.
\newblock \emph{arXiv preprint arXiv:2502.06781}, 2025.

\bibitem[Miao et~al.(2024)Miao, Wang, Wang, Yang, Wang, Qiu, and
  Liu]{miao2024training}
Zichen Miao, Jiang Wang, Ze~Wang, Zhengyuan Yang, Lijuan Wang, Qiang Qiu, and
  Zicheng Liu.
\newblock Training diffusion models towards diverse image generation with
  reinforcement learning.
\newblock In \emph{Proceedings of the IEEE/CVF Conference on Computer Vision
  and Pattern Recognition}, pages 10844--10853, 2024.

\bibitem[Nika et~al.(2024)Nika, Mandal, Kamalaruban, Tzannetos, Radanovi{\'c},
  and Singla]{nika2024reward}
Andi Nika, Debmalya Mandal, Parameswaran Kamalaruban, Georgios Tzannetos, Goran
  Radanovi{\'c}, and Adish Singla.
\newblock Reward model learning vs. direct policy optimization: A comparative
  analysis of learning from human preferences.
\newblock \emph{arXiv preprint arXiv:2403.01857}, 2024.

\bibitem[Podell et~al.(2023)Podell, English, Lacey, Blattmann, Dockhorn,
  M{\"u}ller, Penna, and Rombach]{podell2023sdxl}
Dustin Podell, Zion English, Kyle Lacey, Andreas Blattmann, Tim Dockhorn, Jonas
  M{\"u}ller, Joe Penna, and Robin Rombach.
\newblock Sdxl: Improving latent diffusion models for high-resolution image
  synthesis.
\newblock \emph{arXiv preprint arXiv:2307.01952}, 2023.

\bibitem[Pooladian et~al.(2023)Pooladian, Ben-Hamu, Domingo-Enrich, Amos,
  Lipman, and Chen]{pooladian2023multisample}
Aram-Alexandre Pooladian, Heli Ben-Hamu, Carles Domingo-Enrich, Brandon Amos,
  Yaron Lipman, and Ricky~TQ Chen.
\newblock Multisample flow matching: Straightening flows with minibatch
  couplings.
\newblock \emph{arXiv preprint arXiv:2304.14772}, 2023.

\bibitem[Rafailov et~al.(2023)Rafailov, Sharma, Mitchell, Manning, Ermon, and
  Finn]{rafailov2023direct}
Rafael Rafailov, Archit Sharma, Eric Mitchell, Christopher~D Manning, Stefano
  Ermon, and Chelsea Finn.
\newblock Direct preference optimization: Your language model is secretly a
  reward model.
\newblock \emph{Advances in Neural Information Processing Systems},
  36:\penalty0 53728--53741, 2023.

\bibitem[Rombach et~al.(2022)Rombach, Blattmann, Lorenz, Esser, and
  Ommer]{rombach2022high}
Robin Rombach, Andreas Blattmann, Dominik Lorenz, Patrick Esser, and Bj{\"o}rn
  Ommer.
\newblock High-resolution image synthesis with latent diffusion models.
\newblock In \emph{Proceedings of the IEEE/CVF conference on computer vision
  and pattern recognition}, pages 10684--10695, 2022.

\bibitem[Schuhmann(2022)]{schuhmann2022laion}
Chrisoph Schuhmann.
\newblock Laion aesthetics, Aug 2022.
\newblock URL \url{https://laion.ai/blog/laion-aesthetics/}.

\bibitem[Schulman et~al.(2017)Schulman, Wolski, Dhariwal, Radford, and
  Klimov]{schulman2017proximal}
John Schulman, Filip Wolski, Prafulla Dhariwal, Alec Radford, and Oleg Klimov.
\newblock Proximal policy optimization algorithms.
\newblock \emph{arXiv preprint arXiv:1707.06347}, 2017.

\bibitem[Seo et~al.(2019)Seo, Vecchietti, Lee, and Har]{seo2019rewards}
Minah Seo, Luiz~Felipe Vecchietti, Sangkeum Lee, and Dongsoo Har.
\newblock Rewards prediction-based credit assignment for reinforcement learning
  with sparse binary rewards.
\newblock \emph{IEEE Access}, 7:\penalty0 118776--118791, 2019.

\bibitem[Silver et~al.(2014)Silver, Lever, Heess, Degris, Wierstra, and
  Riedmiller]{silver2014deterministic}
David Silver, Guy Lever, Nicolas Heess, Thomas Degris, Daan Wierstra, and
  Martin Riedmiller.
\newblock Deterministic policy gradient algorithms.
\newblock In \emph{International conference on machine learning}, pages
  387--395. Pmlr, 2014.

\bibitem[Sohl-Dickstein et~al.(2015)Sohl-Dickstein, Weiss, Maheswaranathan, and
  Ganguli]{sohl2015deep}
Jascha Sohl-Dickstein, Eric Weiss, Niru Maheswaranathan, and Surya Ganguli.
\newblock Deep unsupervised learning using nonequilibrium thermodynamics.
\newblock In \emph{International conference on machine learning}, pages
  2256--2265. pmlr, 2015.

\bibitem[Song et~al.(2020{\natexlab{a}})Song, Meng, and
  Ermon]{song2020denoising}
Jiaming Song, Chenlin Meng, and Stefano Ermon.
\newblock Denoising diffusion implicit models.
\newblock \emph{arXiv preprint arXiv:2010.02502}, 2020{\natexlab{a}}.

\bibitem[Song et~al.(2020{\natexlab{b}})Song, Sohl-Dickstein, Kingma, Kumar,
  Ermon, and Poole]{song2020score}
Yang Song, Jascha Sohl-Dickstein, Diederik~P Kingma, Abhishek Kumar, Stefano
  Ermon, and Ben Poole.
\newblock Score-based generative modeling through stochastic differential
  equations.
\newblock \emph{arXiv preprint arXiv:2011.13456}, 2020{\natexlab{b}}.

\bibitem[Sutton et~al.(1998)Sutton, Barto, et~al.]{sutton1998introduction}
Richard~S Sutton, Andrew~G Barto, et~al.
\newblock Introduction to reinforcement learning.
\newblock 1998.

\bibitem[Sutton et~al.(1999)Sutton, McAllester, Singh, and
  Mansour]{sutton1999policy}
Richard~S Sutton, David McAllester, Satinder Singh, and Yishay Mansour.
\newblock Policy gradient methods for reinforcement learning with function
  approximation.
\newblock \emph{Advances in neural information processing systems}, 12, 1999.

\bibitem[Uehara et~al.(2024{\natexlab{a}})Uehara, Zhao, Black, Hajiramezanali,
  Scalia, Diamant, Tseng, Biancalani, and Levine]{uehara2024fine}
Masatoshi Uehara, Yulai Zhao, Kevin Black, Ehsan Hajiramezanali, Gabriele
  Scalia, Nathaniel~Lee Diamant, Alex~M Tseng, Tommaso Biancalani, and Sergey
  Levine.
\newblock Fine-tuning of continuous-time diffusion models as
  entropy-regularized control.
\newblock \emph{arXiv preprint arXiv:2402.15194}, 2024{\natexlab{a}}.

\bibitem[Uehara et~al.(2024{\natexlab{b}})Uehara, Zhao, Black, Hajiramezanali,
  Scalia, Diamant, Tseng, Levine, and Biancalani]{uehara2024feedback}
Masatoshi Uehara, Yulai Zhao, Kevin Black, Ehsan Hajiramezanali, Gabriele
  Scalia, Nathaniel~Lee Diamant, Alex~M Tseng, Sergey Levine, and Tommaso
  Biancalani.
\newblock Feedback efficient online fine-tuning of diffusion models.
\newblock \emph{arXiv preprint arXiv:2402.16359}, 2024{\natexlab{b}}.

\bibitem[Uehara et~al.(2024{\natexlab{c}})Uehara, Zhao, Hajiramezanali, Scalia,
  Eraslan, Lal, Levine, and Biancalani]{uehara2024bridging}
Masatoshi Uehara, Yulai Zhao, Ehsan Hajiramezanali, Gabriele Scalia, Gokcen
  Eraslan, Avantika Lal, Sergey Levine, and Tommaso Biancalani.
\newblock Bridging model-based optimization and generative modeling via
  conservative fine-tuning of diffusion models.
\newblock \emph{Advances in Neural Information Processing Systems},
  37:\penalty0 127511--127535, 2024{\natexlab{c}}.

\bibitem[Uehara et~al.(2025)Uehara, Su, Zhao, Li, Regev, Ji, Levine, and
  Biancalani]{uehara2025reward}
Masatoshi Uehara, Xingyu Su, Yulai Zhao, Xiner Li, Aviv Regev, Shuiwang Ji,
  Sergey Levine, and Tommaso Biancalani.
\newblock Reward-guided iterative refinement in diffusion models at test-time
  with applications to protein and dna design.
\newblock \emph{arXiv preprint arXiv:2502.14944}, 2025.

\bibitem[Wallace et~al.(2024)Wallace, Dang, Rafailov, Zhou, Lou, Purushwalkam,
  Ermon, Xiong, Joty, and Naik]{wallace2024diffusion}
Bram Wallace, Meihua Dang, Rafael Rafailov, Linqi Zhou, Aaron Lou, Senthil
  Purushwalkam, Stefano Ermon, Caiming Xiong, Shafiq Joty, and Nikhil Naik.
\newblock Diffusion model alignment using direct preference optimization.
\newblock In \emph{Proceedings of the IEEE/CVF Conference on Computer Vision
  and Pattern Recognition}, pages 8228--8238, 2024.

\bibitem[Wang et~al.(2023)Wang, Li, Shao, Xu, Dai, Li, Chen, Wu, and
  Sui]{wang2023math}
Peiyi Wang, Lei Li, Zhihong Shao, RX~Xu, Damai Dai, Yifei Li, Deli Chen, Yu~Wu,
  and Zhifang Sui.
\newblock Math-shepherd: Verify and reinforce llms step-by-step without human
  annotations.
\newblock \emph{arXiv preprint arXiv:2312.08935}, 2023.

\bibitem[Wang et~al.(2024)Wang, Zheng, Ye, Xue, Huang, and
  Gu]{wang2024diffusion}
Xinyou Wang, Zaixiang Zheng, Fei Ye, Dongyu Xue, Shujian Huang, and Quanquan
  Gu.
\newblock Diffusion language models are versatile protein learners.
\newblock \emph{arXiv preprint arXiv:2402.18567}, 2024.

\bibitem[Watson et~al.(2023)Watson, Juergens, Bennett, Trippe, Yim, Eisenach,
  Ahern, Borst, Ragotte, Milles, et~al.]{watson2023novo}
Joseph~L Watson, David Juergens, Nathaniel~R Bennett, Brian~L Trippe, Jason
  Yim, Helen~E Eisenach, Woody Ahern, Andrew~J Borst, Robert~J Ragotte, Lukas~F
  Milles, et~al.
\newblock De novo design of protein structure and function with rfdiffusion.
\newblock \emph{Nature}, 620\penalty0 (7976):\penalty0 1089--1100, 2023.

\bibitem[Wu et~al.(2023)Wu, Hao, Sun, Chen, Zhu, Zhao, and
  Li]{wu2023humanpreferencescorev2}
Xiaoshi Wu, Yiming Hao, Keqiang Sun, Yixiong Chen, Feng Zhu, Rui Zhao, and
  Hongsheng Li.
\newblock Human preference score v2: A solid benchmark for evaluating human
  preferences of text-to-image synthesis, 2023.

\bibitem[Xing et~al.(2025)Xing, Saha, He, Hao, Vicol, Ryu, Li, Singla, Young,
  Li, et~al.]{xing2025focus}
Xiaoying Xing, Avinab Saha, Junfeng He, Susan Hao, Paul Vicol, Moonkyung Ryu,
  Gang Li, Sahil Singla, Sarah Young, Yinxiao Li, et~al.
\newblock Focus-n-fix: Region-aware fine-tuning for text-to-image generation.
\newblock \emph{arXiv preprint arXiv:2501.06481}, 2025.

\bibitem[Xu et~al.(2023)Xu, Liu, Wu, Tong, Li, Ding, Tang, and
  Dong]{xu2023imagereward}
Jiazheng Xu, Xiao Liu, Yuchen Wu, Yuxuan Tong, Qinkai Li, Ming Ding, Jie Tang,
  and Yuxiao Dong.
\newblock Imagereward: learning and evaluating human preferences for
  text-to-image generation.
\newblock In \emph{Proceedings of the 37th International Conference on Neural
  Information Processing Systems}, pages 15903--15935, 2023.

\bibitem[Xu et~al.(2024)Xu, Huang, Cheng, Yang, Xu, Wang, Duan, Yang, Jin, Li,
  et~al.]{xu2024visionreward}
Jiazheng Xu, Yu~Huang, Jiale Cheng, Yuanming Yang, Jiajun Xu, Yuan Wang, Wenbo
  Duan, Shen Yang, Qunlin Jin, Shurun Li, et~al.
\newblock Visionreward: Fine-grained multi-dimensional human preference
  learning for image and video generation.
\newblock \emph{arXiv preprint arXiv:2412.21059}, 2024.

\bibitem[Yim et~al.(2023)Yim, Trippe, De~Bortoli, Mathieu, Doucet, Barzilay,
  and Jaakkola]{yim2023se}
Jason Yim, Brian~L Trippe, Valentin De~Bortoli, Emile Mathieu, Arnaud Doucet,
  Regina Barzilay, and Tommi Jaakkola.
\newblock Se (3) diffusion model with application to protein backbone
  generation.
\newblock \emph{arXiv preprint arXiv:2302.02277}, 2023.

\bibitem[Zhang and Xu(2023)]{zhang2023towards}
Hengtong Zhang and Tingyang Xu.
\newblock Towards controllable diffusion models via reward-guided exploration.
\newblock \emph{arXiv preprint arXiv:2304.07132}, 2023.

\bibitem[Zhang et~al.(2024{\natexlab{a}})Zhang, Tzeng, Du, and
  Kislyuk]{zhang2024large}
Yinan Zhang, Eric Tzeng, Yilun Du, and Dmitry Kislyuk.
\newblock Large-scale reinforcement learning for diffusion models.
\newblock In \emph{European Conference on Computer Vision}, pages 1--17.
  Springer, 2024{\natexlab{a}}.

\bibitem[Zhang et~al.(2024{\natexlab{b}})Zhang, Shen, Zhang, Ye, Luo, Shi, Du,
  and Tao]{zhang2024aligning}
Ziyi Zhang, Li~Shen, Sen Zhang, Deheng Ye, Yong Luo, Miaojing Shi, Bo~Du, and
  Dacheng Tao.
\newblock Aligning few-step diffusion models with dense reward difference
  learning.
\newblock \emph{arXiv preprint arXiv:2411.11727}, 2024{\natexlab{b}}.

\bibitem[Zhang et~al.(2024{\natexlab{c}})Zhang, Zhang, Zhan, Luo, Wen, and
  Tao]{zhang2024confronting}
Ziyi Zhang, Sen Zhang, Yibing Zhan, Yong Luo, Yonggang Wen, and Dacheng Tao.
\newblock Confronting reward overoptimization for diffusion models: A
  perspective of inductive and primacy biases.
\newblock \emph{arXiv preprint arXiv:2402.08552}, 2024{\natexlab{c}}.

\bibitem[Zhao et~al.(2024)Zhao, Uehara, Scalia, Kung, Biancalani, Levine, and
  Hajiramezanali]{zhao2024adding}
Yulai Zhao, Masatoshi Uehara, Gabriele Scalia, Sunyuan Kung, Tommaso
  Biancalani, Sergey Levine, and Ehsan Hajiramezanali.
\newblock Adding conditional control to diffusion models with reinforcement
  learning.
\newblock \emph{arXiv preprint arXiv:2406.12120}, 2024.

\bibitem[Zhuang et~al.(2023)Zhuang, Lei, Liu, Wang, and
  Guo]{zhuang2023behavior}
Zifeng Zhuang, Kun Lei, Jinxin Liu, Donglin Wang, and Yilang Guo.
\newblock Behavior proximal policy optimization.
\newblock \emph{arXiv preprint arXiv:2302.11312}, 2023.

\end{thebibliography}
\bibliographystyle{plainnat}

\newpage
\appendix
\section{Appendix / supplemental material}



\subsection{Preliminaries on protein backbone generation}
\label{preliminary_protein}
\subsubsection{Protein backbone parameterization}
The parameterization of a protein backbone for generative modeling commonly follows principles similar to those used in AlphaFold2~\citep{jumper2021highly}. A protein is a sequence of amino acid residues, and the backbone of each residue $n$ is primarily defined by four heavy atoms: the amide nitrogen $N_n$, the alpha-carbon $C_{\alpha,n}$, the carbonyl carbon $C_n$, and the oxygen $O_n$.

To describe the coordinates of each heavy atom, each residue $n$ is associated with a local coordinate frame, represented as a rigid transformation $T_n \in \SE(3)$ , where $\SE(3)$ is the Special Euclidean group of three-dimensional rotations and translations. 
For a protein with N residues, the complete set of backbone frames is collectively denoted as $T=[T_1, ... ,T_N] \in \SE(3)_N$.
Once these frames are established, the global Cartesian coordinates of the N$_n$, $C_{\alpha,n}$, and C$_n$ atoms for each residue are determined by applying the respective transformation $T_n$ to a set of idealized, canonical local coordinates for these atoms. 
The position of the oxygen atom O$_n$ can be finalized using an additional torsion angle that orients it correctly relative to the $C_n$ - $C_{\alpha,n}$ segment.

\subsubsection{Flow matching on $\SE(3)$}

A natural approach to define flows on $\SE(3)$ for tasks like protein backbone generation is to first decompose $\SE(3)$ into its rotational component $\SO(3)$ (Special Orthogonal group) and its translational component $\mathbb{R}^3$. This decomposition then allows for the construction of independent flows on $\SO(3)$ and $\mathbb{R}^3$. Given that flow matching in Euclidean spaces $\mathbb{R}^3$ is a well-studied area~\citep{liu2022flow, lipman2022flow}, we will primarily focus our discussion here on the formulation of flow matching on the $\SO(3)$ manifold.

Let $\rho_0$ and $\rho_1$ be two probability density functions on $\SO(3)$, representing  a source distribution (\eg, an easy-to-sample prior) and a target data distribution (\eg, observed protein backbone frame orientations), respectively.
A flow is defined by a time-dependent diffeomorphism $\psi_t: \SO(3) \to \SO(3)$ for $t \in [0,1]$. This flow $\psi_t$ is the solution to an ordinary differential equation (ODE):
\begin{equation}
    \frac{d}{dt}\psi_t(R) = u_t(\psi_t(R)),
    \label{eq:ode_flow}
\end{equation}
with the initial condition $\psi_0(R) = R$ (if flowing from $R$ at $t=0$), where $R \in \SO(3)$ and $u_t: [0,1] \times \SO(3) \to \mathcal{T}\SO(3)$ is a time-dependent smooth vector field on $\SO(3)$, where $\mathcal{T}\SO(3)$ means the tangent space at $\SO(3)$. The flow $\psi_t$ generates the probability path $\rho_t$ if it pushes forward the source distribution, \eg, $\rho_t = [\psi_t]_{\#}\rho_0$.

Conditional flow matching (CFM) aims to regress a parameterized vector field $v_\theta(t, R_t)$ to a target conditional vector field $u_t(R_t | R_0, R_1)$ that transports points between $R_0 \sim \rho_0$ and $R_1 \sim \rho_1$. A common strategy is to construct this conditional flow along the geodesic path. For $SO(3)$, the geodesic interpolant $R_t$ from $R_0$ to $R_1$ is $R_t = \text{exp}_{R_0}(t \cdot \text{log}_{R_0}(R_1))$. The corresponding target conditional vector field $u_t(R_t | R_0, R_1)$ is its time derivative.
If we define the flow as transforming samples from $\rho_0$ (\eg, noise at $t=0$) to $\rho_1$ (data at $t=1$), a common expression for the target vector field, representing the velocity at $R_t$ pointing towards $R_1$, is given by~\citep{bose2023se}: 
\begin{equation}
    u_t(R_t | R_0, R_1) = \frac{\text{log}_{R_t}(R_1)}{t},
    \label{eq:ut_so3}
\end{equation}
where $\text{log}_{R_t}(R_1)$ is an element of the Lie algebra $\mathfrak{so}(3)$ in the tangent space at $R_t$.
Given these components, the flow matching objective for $SO(3)$ can be formulated as:
\begin{equation}
    \mathcal{L}_{SO(3)}(\theta) = \mathbb{E}_{t \sim \mathcal{U}(0,1), q(R_0, R_1), R_t \sim \rho_t(R_t|R_0,R_1)} \left\| v_\theta(t, R_t) - u_t(R_t|R_0,R_1) \right\|^2_{SO(3)}.
    \label{eq:loss_so3}
\end{equation}
Here, $t$ is sampled uniformly from $\mathcal{U}(0,1)$, $q(R_0, R_1)$ is a coupling of samples from the source and target distributions, and $\rho_t(R_t|R_0,R_1)$ is the conditional probability of $R_t$ given $R_0$ and $R_1$. 
For the coupling $q(R_0, R_1)$, samples $R_0 \sim \rho_0$ and $R_1 \sim \rho_1$ can be drawn independently from their respective distributions. Alternatively, to achieve straighter interpolation paths and improved training stability, many approaches approximate the optimal transport (OT) coupling $\pi(R_0, R_1)$ using minibatch techniques~\citep{pooladian2023multisample}.

\subsection{Proof of Lemma 1}

Given two diffusion models with Gaussian conditional distributions:
\begin{align}
    p_\theta(\mathbf{x}_{T-t}|\mathbf{x}_{T-t+1}, \mathbf{c}) &= \mathcal{N}(\mu_\theta, \sigma^2 I), \\
    p_{\theta^0}(\mathbf{x}^0_{T-t}|\mathbf{x}_{T-t+1}, \mathbf{c}) &= \mathcal{N}(\mu_{\theta^0}, \sigma^2 I).
\end{align}
For Gaussian distributions, the gradient of KL divergence simplifies to:
\begin{align}
    \nabla_\theta D_{KL}(p_\theta \| p_{\theta^0}) = \nabla_\theta \frac{1}{2\sigma^2} \|\mu_\theta - \mu_{\theta^0}\|^2.
\end{align}

The expectation over sample pairs $\mathbf{x}_{T-t} \sim p_\theta$ and $\mathbf{x}^0_{T-t} \sim p_{\theta^0}$ is:
\begin{align}
    \mathbb{E}_{\mathbf{x}_{T-t} \sim p_\theta,\mathbf{x}^0_{T-t} \sim p_{\theta^0} }\left[ (\mathbf{x}_{T-t} - \mathbf{x}^0_{T-t})^2 \right] &= \mathbb{E}_{\epsilon_\theta, \epsilon_{\theta^0}\sim\mathcal{N}(0, \sigma^2 I)}\left[ (\mu_\theta - \mu_{\theta^0} + \epsilon_\theta - \epsilon_{\theta^0})^2 \right] \nonumber \\
    &= \|\mu_\theta - \mu_{\theta^0}\|^2 + 2 (\mu_\theta - \mu_{\theta^0})\mathbb{E}\left[\epsilon_{\theta}-\epsilon_{\theta^0}   \right]  \nonumber
    \\ &\quad  + \mathbb{E}\left[ \|\epsilon_\theta\|^2 \right] - 2\mathbb{E}\left[ \epsilon_\theta \epsilon_{\theta^0}\right]+ \mathbb{E}\left[ \|\epsilon_{\theta^0}\|^2 \right]\nonumber \\
    &= \|\mu_\theta - \mu_{\theta^0}\|^2 + 2\sigma^2d,
\end{align}
where $\epsilon_\theta, \epsilon_{\theta^0} \sim \mathcal{N}(0, \sigma^2 I)$ are independent noise terms and $d$ is the state dimension.

Substituting the KL expression into the MSE yields the lemma statement:
\begin{align}
   \nabla_\theta \mathbb{E}\left[ (\mathbf{x}_{T-t} - \mathbf{x}^0_{T-t})^2 \right] = 2\sigma^2 \nabla_\theta D_{KL}(p_\theta \| p_{\theta^0}).
\end{align}


\subsection{Value function architectures}

The learned value function serves as a crucial component in VARD, providing the supervision signal for fine-tuning. We now specify the architectures used to implement this value function in our different experimental settings.

\textbf{Non-differentiable rewards.}
For experiments on protein secondary structure diversity, we employ the GVP-Transformer encoder~\citep{hsu2022learning} to generate translation-invariant and rotation-equivariant representations. This encoder uses an embedding dimension of 512 and 4 layers, followed by a 2-layer MLP head that predicts helix and strand percentages. The complete architecture comprises approximately 42M parameters. For experiments on compressibility and incompressibility, we train a standard ResNet-18~\citep{he2016deep} (11M parameters) with a regression head. This model is trained to predict the compressed file size for the data state at each diffusion timestep, serving directly as our differentiable value function.

\textbf{Differentiable rewards.}
The differentiable rewards utilized in our experiments consist of four models designed for scoring human preferences, namely: Aesthetic Scorer, PickScore, ImageReward, and HPSv2. In our preliminary experiments, we initially employed a value function architecture based on the DPOK approach~\citep{fan2023dpok}; this is typically a simple MLP whose primary role is to stabilize the training process.
However, we found this architecture to be suboptimal for accurately predicting values from intermediate diffusion states, largely due to the MLP's limited capacity to model the complex and nuanced distribution underlying human preference scores.
Consequently, we adopted a more effective strategy by directly adapting these differentiable reward models themselves to function as the value function, using Low-Rank Adaptation (LoRA)~\citep{hu2022lora}. 
Specifically, we integrated LoRA with an alpha of 32 and a dropout rate of 0.1 into both the self-attention mechanisms and the subsequent feed-forward network layers of the visual transformer encoders within these reward models. 
This LoRA-based implementation offers distinct advantages.
Firstly, its architecture, being derived from the reward models themselves, is inherently more suitable for modeling complex human preferences compared to a simple MLP.
Secondly, despite the large parameter from the original reward models, LoRA ensures training efficiency for the value function by updating only a small fraction of these parameters, drastically reducing the GPU memory during training.
Finally, leveraging the pre-trained weights of these reward models provides a strong initialization that significantly enhances the value function's learning effectiveness; we found that the value function learns rapidly and performs well with only a remarkably small number of sampled trajectories.

\subsection{Implementation details}
%
% ...
For both protein structure and image generation, we employ a 50-step diffusion sampling process, utilizing the official FoldFlow implementation\footnote{https://github.com/DreamFold/FoldFlow} for proteins and a DDPM sampler~\citep{ho2020denoising} for images. 
Notably, our proposed method also supports DDIM~\citep{song2020denoising} as an alternative sampler.
In the context of protein generation, we observe that the protein's secondary structure is largely determined during the early diffusion stages, while subsequent timesteps primarily involve minor adjustments to atomic coordinates. Consequently, for protein-related objectives, the FoldFlow-2 network is trained across all 50 diffusion timesteps.
Conversely, for image generation, we find that objectives such as human preferences (\eg, aesthetic scores) and compressibility are more strongly influenced by fine-grained details that are finalized in the later stages of the diffusion process. Therefore, to achieve a favorable trade-off between computational efficiency and the quality of preference alignment, the U-Net is fine-tuned only on the final 10 diffusion steps for these image-specific tasks.

All experiments are conducted on 8 H800 GPUs, taking approximately 30 minutes for training, and utilize the AdamW optimizer~\citep{loshchilov2017decoupled}, configured with $\beta_1=0.9$, $\beta_2=0.999$, $\epsilon=1e-8$, and a weight decay of 0.01. 
Additionally, we apply gradient clipping with a maximum norm of 1.0.
The subsequent paragraph outlines the detailed configurations for each experiment.

\begin{table}[t]
\centering
\caption{Hyperparameters for non-differentiable rewards.}
\label{tab:non_differentiable_configs}
\setlength{\tabcolsep}{6pt} 
\begin{tabular*}{\textwidth}{@{\extracolsep{\fill}}l|cc} 
\toprule
\textbf{Hyperparameters} & \textbf{Protein secondary} & \textbf{Image compressibility} \\
& \textbf{structure diversity} & \textbf{and incompressibility} \\
\midrule
\multicolumn{3}{l}{\textit{Value Function (PRM) Pretraining}} \\
\midrule
Learning rate & 3e-4 & 1e-4 \\
Global batch size & 128 & 32 \\
Train steps & 2k & 3k \\
\midrule
\multicolumn{3}{l}{\textit{Diffusion Model Fine-tuning (with VARD)}} \\
\midrule
KL regularization weight $\eta$ & 0.1 & 1 \\
Base model learning rate & 5e-5 &  1e-6 \\
Value function learning rate & 1e-6 & 1e-5 \\
Gradient accumulate steps & 50 & 2 \\
Train steps & 30 & 200 \\
Global batch size & 16 & 32 \\
\bottomrule
\end{tabular*} 
\end{table}

\begin{table}[t]
\centering
\caption{Hyperparameters for differentiable rewards.}
\label{tab:differentiable_configs}
\setlength{\tabcolsep}{6pt} 
\begin{tabular*}{\textwidth}{@{\extracolsep{\fill}}l|ccc} 
\toprule
\textbf{Hyperparameters} & \textbf{Aesthetic Score} & \textbf{PickScore} & \textbf{ImageReward} \\
\midrule
\multicolumn{4}{l}{\textit{Value Function (PRM) Pretraining}} \\
\midrule
Learning rate & 1e-6  &  5e-6 & 1e-5 \\
Batch size & 32 & 32 & 32\\
Train steps  & 8 & 8 & 8\\
\midrule
\multicolumn{4}{l}{\textit{Diffusion Model Fine-tuning (with VARD)}} \\
\midrule
KL regularization weight $\eta$ & 100 &  0.5 & 20\\
Base model learning rate &  1e-6 & 5e-6  & 1e-5 \\
Value function learning rate & 5e-6 & 5e-6  & 5e-6\\
Gradient accumulate steps & 2 & 2 &2\\
Train steps & 1k & 1k & 1k\\
Global batch size & 32 & 32 & 32\\
\bottomrule
\end{tabular*} 
\end{table}

\textbf{Non-differentiable rewards.}  
For the prediction of protein secondary structure, our value function is 
\begin{wraptable}{r}{0.45\textwidth}
    \centering
    \caption{Impact of hyperparameter $w_b$ on the performance of FoldFlow-2-ReFT relative to FoldFlow-2-base. While the $w_b=2$ setting leads to FoldFlow-2-ReFT underperforming its baseline, our adjusted value of $w_b=5$ significantly improves its reward.}
    \label{tab:protein_weight}
    \begin{tabular}{ccc}
        \toprule
        $w_b$ & Model & Rewards \\
        \midrule
        2   & FoldFlow-2-base  & 0.83\\
        2   & FoldFlow-2-ReFT  & 0.76\\
        5   & FoldFlow-2-base  & 0.85\\
        5   & FoldFlow-2-ReFT  & 1.27\\
        \bottomrule
    \end{tabular}
\end{wraptable}
trained on the SCOPe dataset~\cite{fox2014scope} to bypass data generation via FoldFlow-2 and thereby expedite the training process.
To predict the compressed size of the image, we train the value function with images generated by Stable Diffusion 1.4.
We provide the details of hyperparameters in Table~\ref{tab:non_differentiable_configs}. 
Besides, for experiments with protein generation, careful readers may find that in the original FoldFlow-2 paper~\citep{bose2023se}, $w_b$ is set to be 2 instead of 5. However, we find that with the setting of $w_a=1$, $w_b=2$ and $w_c=0.5$, the reward of proteins that generate from the checkpoint of FoldFlow-2-ReFT is even lower than the FoldFlow-2-base, where we refer to the FoldFlow-2-ReFT as FoldFlow-2-base after the reinforced fine-tuning introduced in the original FoldFlow-2 paper, as shown in Table~\ref{tab:protein_weight}. We hypothesize that this is a typo in the original paper. So we increase the $w_b$ to 5, so that the reward of FoldFlow-2-ReFT can be higher that FoldFlow-2.

\textbf{Differentiable rewards.} Hyperparameters are presented in Table~\ref{tab:differentiable_configs}.

\textbf{Unseen prompt generalization.} As referenced in the main text, Table~\ref{table:cross_prompts} includes metrics from the DRaFT~\citep{clark2023directly} and PRDP~\citep{Deng_2024_CVPR} papers. To provide full context for these comparisons, we detail relevant aspects of their experimental settings here. Notably, DDPO and PRDP are trained using 512,000 sampled images. In contrast, other methods such as ReFL, DRaFT-1, and DRaFT-LV—which are reward backpropagation approaches—are trained on 160,000 sampled images. Our method (VARD) also follows this 160,000-sample regimen to maintain parity, particularly with these reward backpropagation techniques.

\textbf{Cross-reward validation and prior consistency-reward balance.} For Aesthetic Score, we use prompts from the animal category and generate images from 10 random seeds. For PickScore and ImageReward, we sample 100 prompts from the Pick-a-Pic and ImageRewardDB datasets, respectively, and generate images on 5 seeds.

\subsection{Additional experimental results}
\label{sec:ablation_study}

\subsubsection{Effect of KL regularization.}
\begin{figure}[t]
\centering
\includegraphics[scale=0.45]{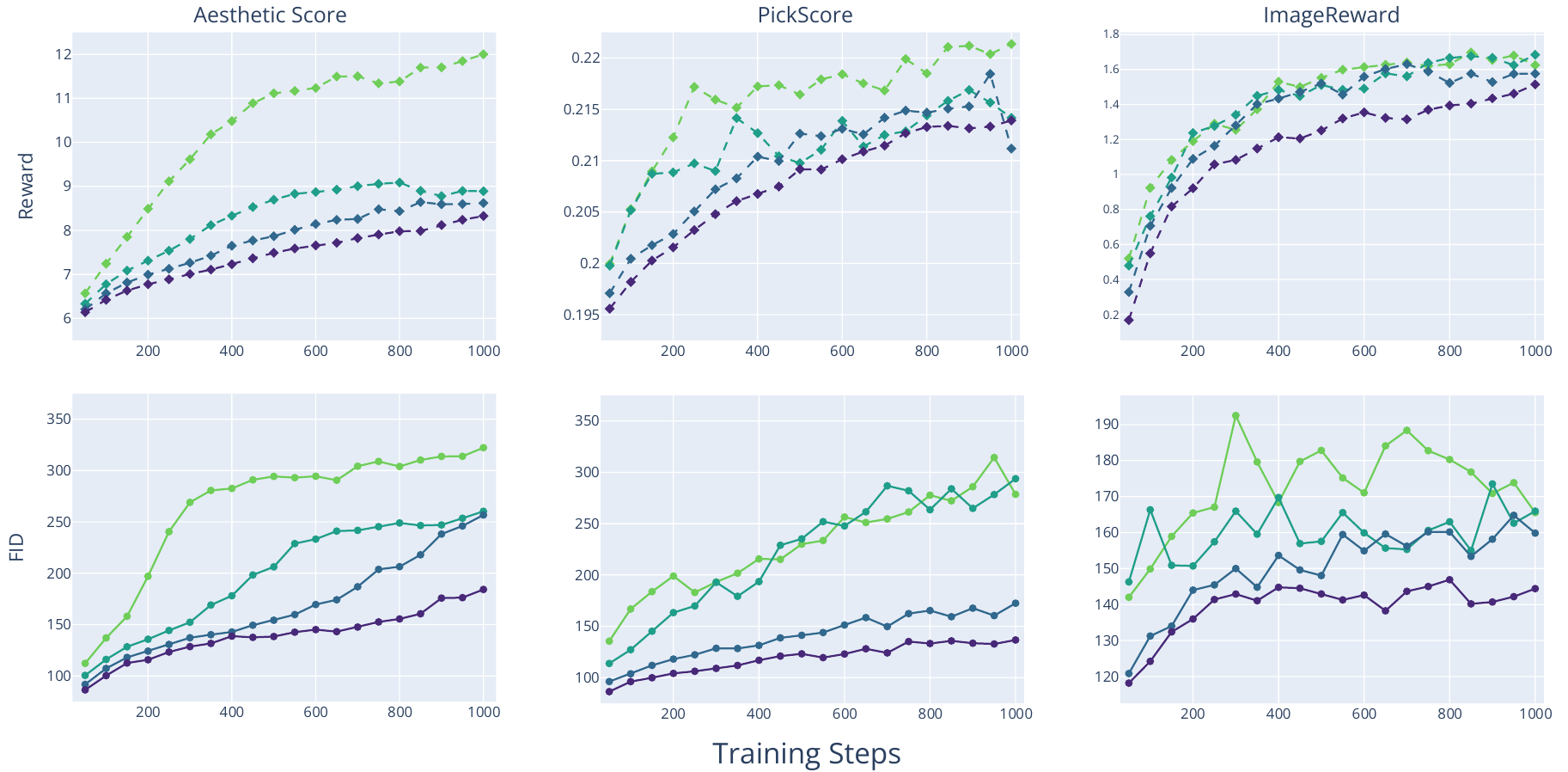}
\vspace{-5mm}
\caption{
\textbf{Training trend of FID (left) and reward (right) under KL regularization.} Color intensity correlates with $\eta$ magnitude (darker corresponds to higher values). Higher $\eta$ configurations maintain lower FID scores (better prior preservation) while exhibiting slower reward growth, quantitatively demonstrating the preservation-optimization tradeoff.
}
\label{figure:ab_eta}
\end{figure}

We systematically adjust the $\eta$ parameter controlling pretrained model regularization strength, using values $[0,20,50,100]$ for Aesthetic Score, $[0,0.1,0.5,1]$ for PickScore, and $[0,1,5,20]$ for ImageReward. Figure~\ref{figure:ab_eta} confirm that higher $\eta$ values: (1) better preserve the pretrained model's priors during training, and (2) proportionally slow reward metric improvement, which demonstrates the inherent trade-off between prior preservation and reward optimization speed.

\subsubsection{Uncurated samples}
\begin{figure}[h]
\centering
\includegraphics[scale=0.33]{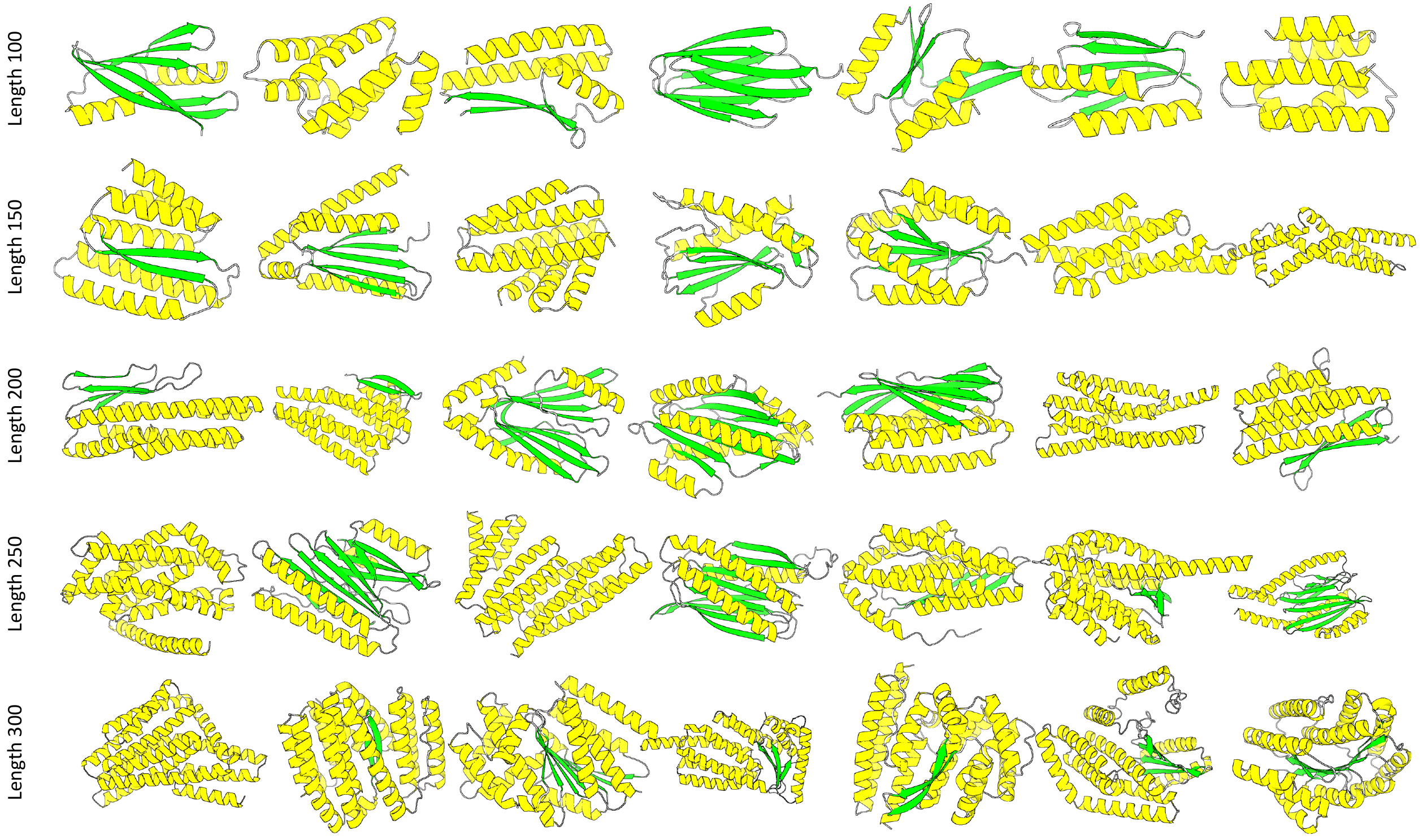}
\caption{\textbf{Uncurated protein structures generated by FoldFlow-2-VARD.} The showcased samples include proteins with various secondary structures, such as those containing beta-sheets. This illustrates that FoldFlow-2-VARD maintains a broad generative capacity, retaining its ability to produce entirely alpha-helical proteins alongside more complex structures.}
\label{figure:appendix_protein_visual}
\end{figure}

\begin{figure}[t]
\centering
\includegraphics[scale=0.40]{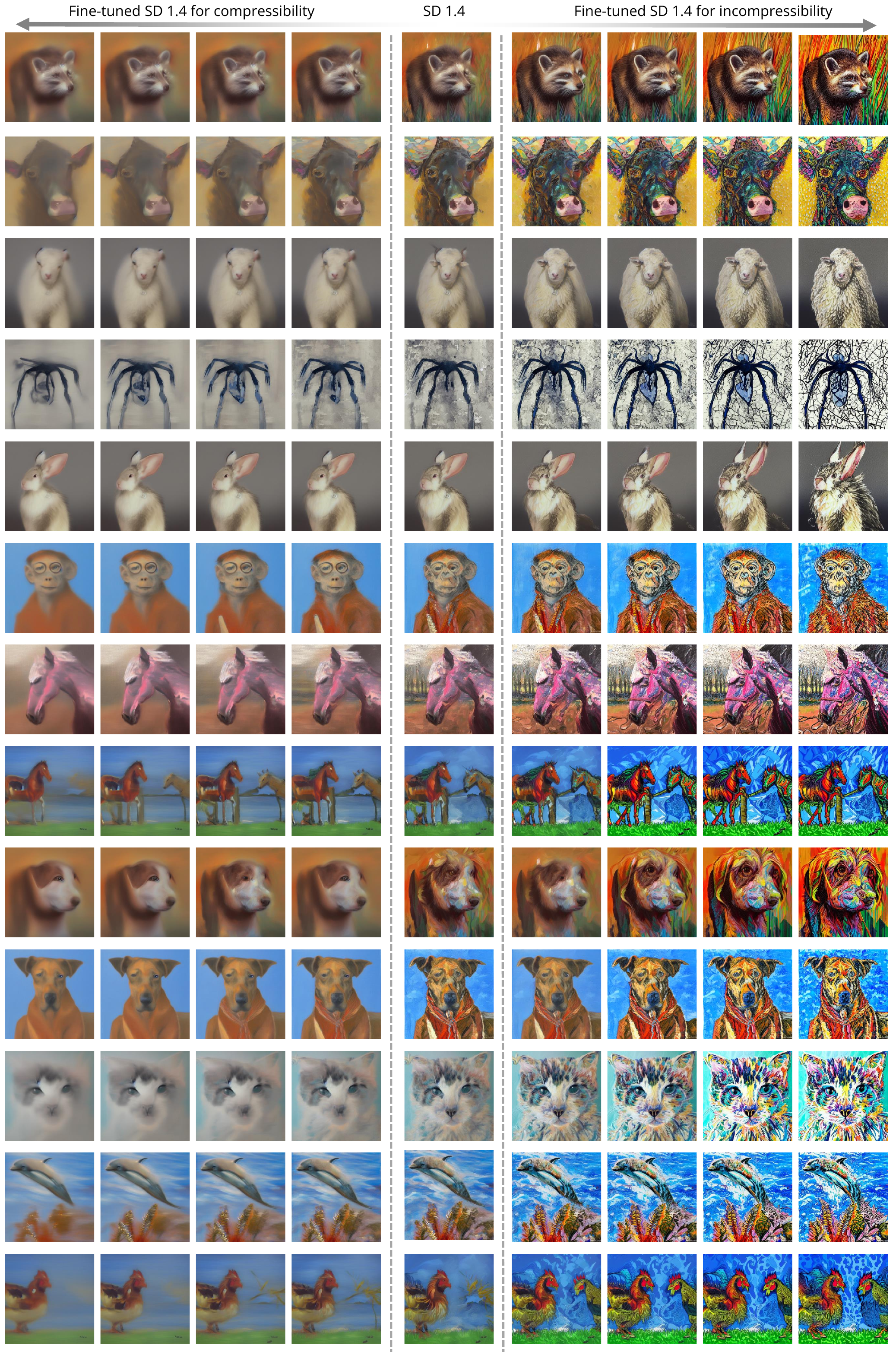}
\caption{\textbf{Uncurated samples of fine-tuning on compressibility and incompressibility as rewards.}}
\label{figure:appendix_compressibility}
\end{figure}

\begin{figure}[t]
\centering
\includegraphics[scale=0.35]{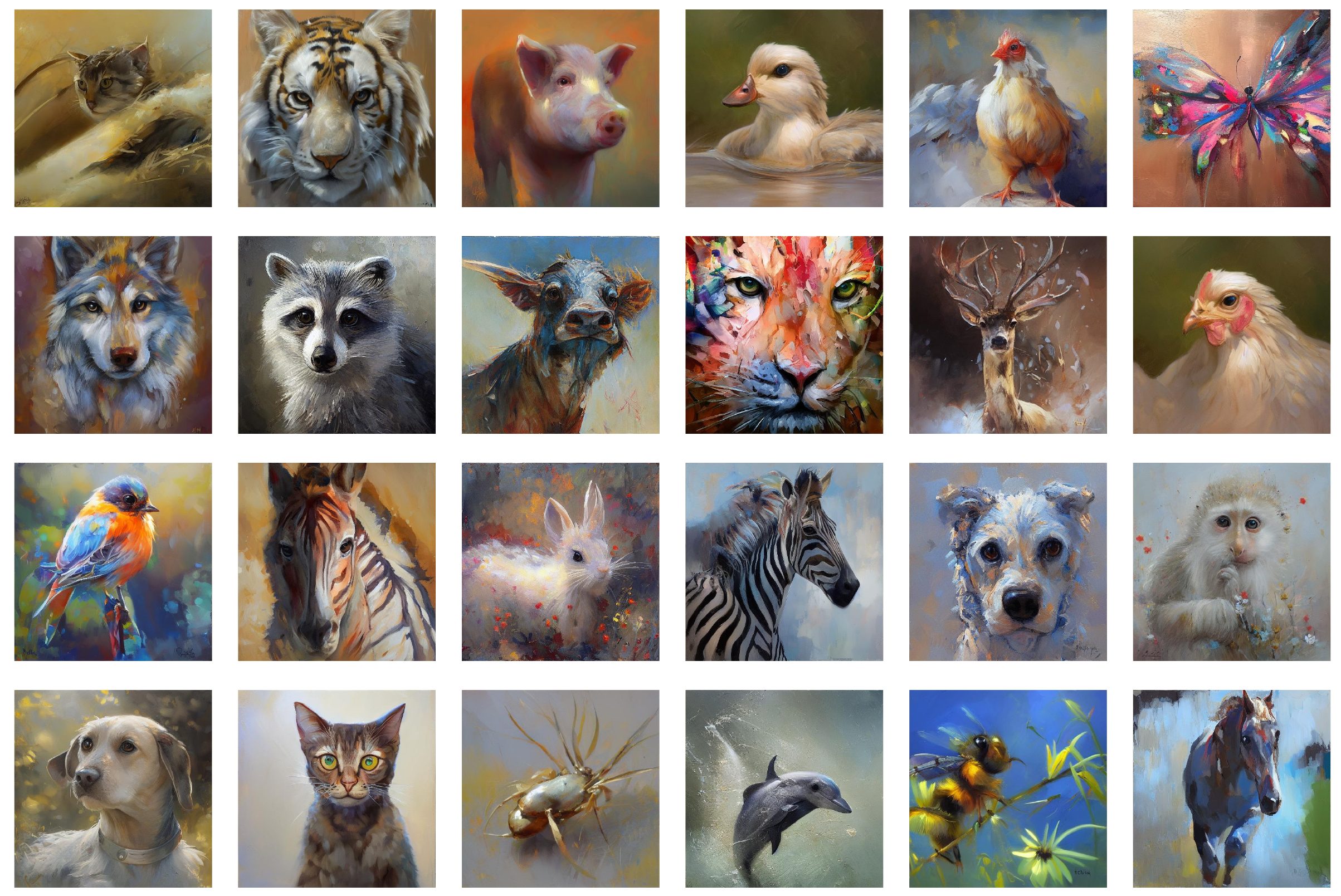}
\caption{\textbf{Uncurated samples of fine-tuning on Aesthetic Score as the reward.}}
\label{figure:appendix_aes}
\end{figure}

\begin{figure}[t]
\centering
\includegraphics[scale=0.33]{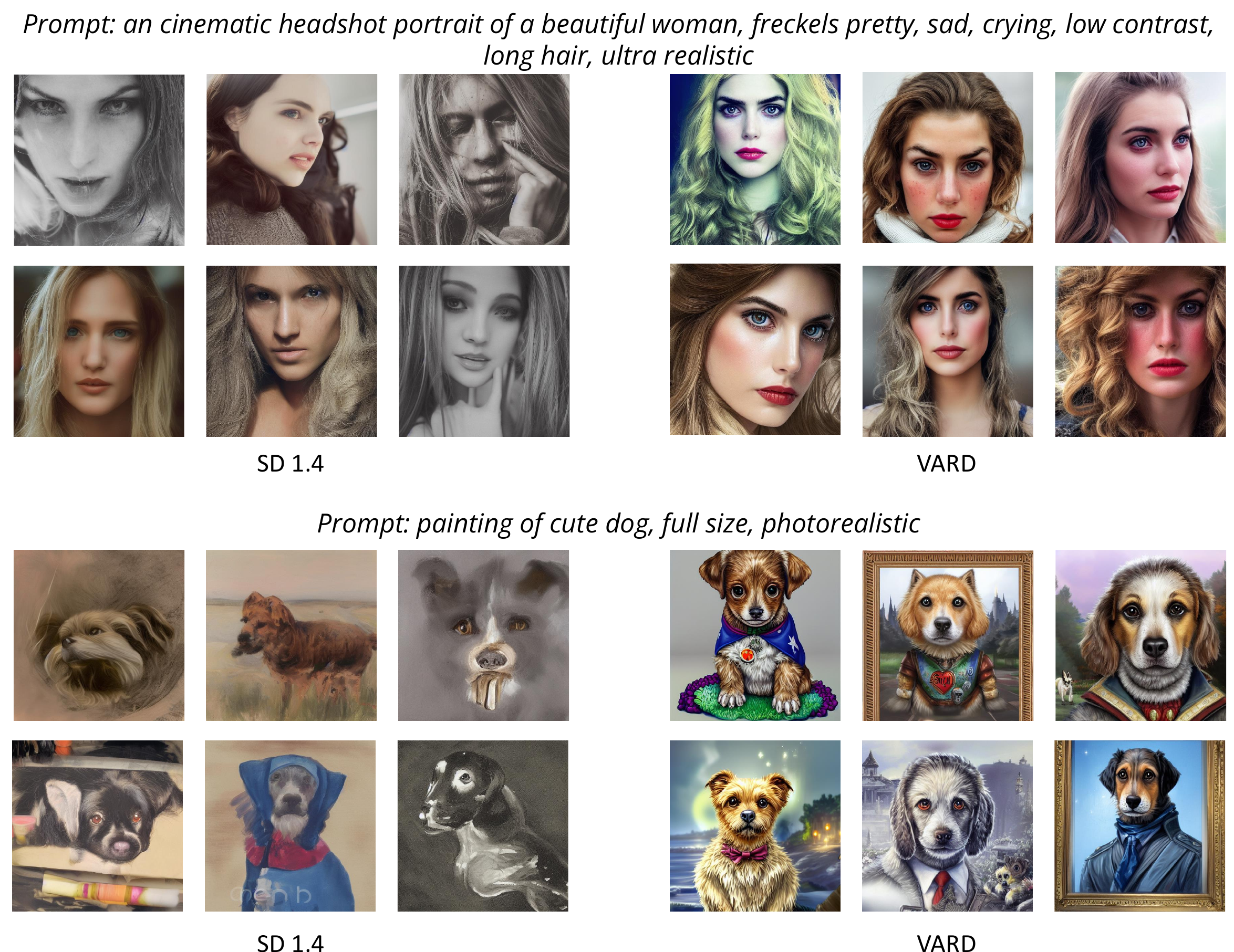}
\caption{\textbf{Further comparison of the base model and VARD fine-tuned version with ImageReward as reward.}}
\label{figure:appendix_imagereward}
\end{figure}

\begin{figure}[t]
\centering
\includegraphics[scale=0.33]{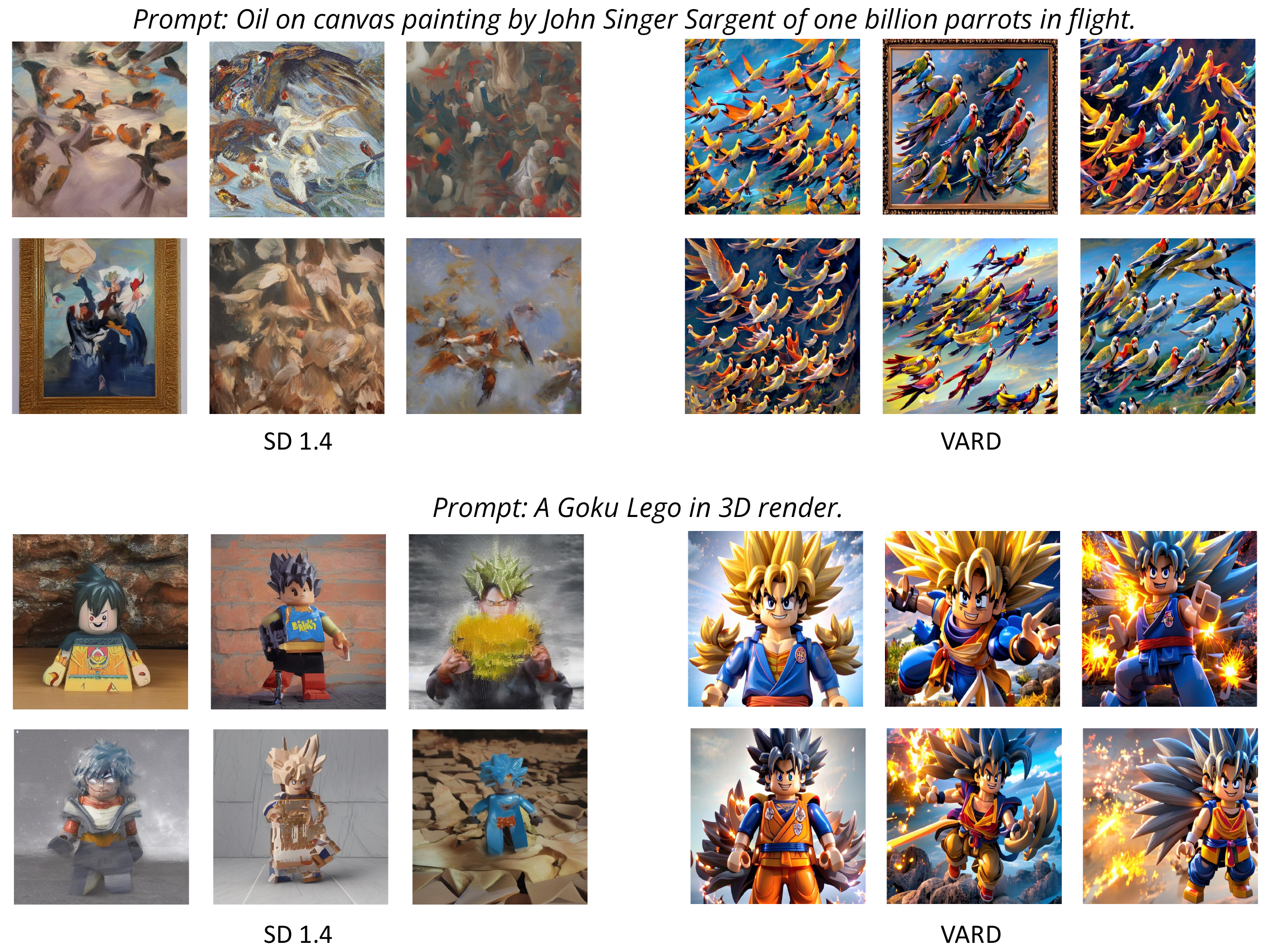}
\caption{\textbf{Further comparison of the base model and VARD fine-tuned version with HPSv2 as reward.}}
\label{figure:appendix_hpsv2}
\end{figure}

Here, we present uncurated samples from models fine-tuned with VARD for various reward functions. These include results for protein secondary structure diversity (Figure~\ref{figure:appendix_protein_visual}), compressibility and incompressibility (Figure~\ref{figure:appendix_compressibility}), Aesthetic Score (Figure~\ref{figure:appendix_aes}), PickScore and HPSv2 (Figure~\ref{figure:appendix_hpsv2}), and ImageReward (Figure~\ref{figure:appendix_imagereward}).


\subsection{Experiments on other diffusion models
}

To further validate the effectiveness of VARD, we extended our evaluations by fine-tuning more recent and advanced base diffusion models, SDXL~\citep{podell2023sdxl}. By testing VARD on such a capable architecture, we aim to demonstrate its adaptability and consistent performance benefits.

\begin{figure}[t]
\centering
\includegraphics[scale=0.33]{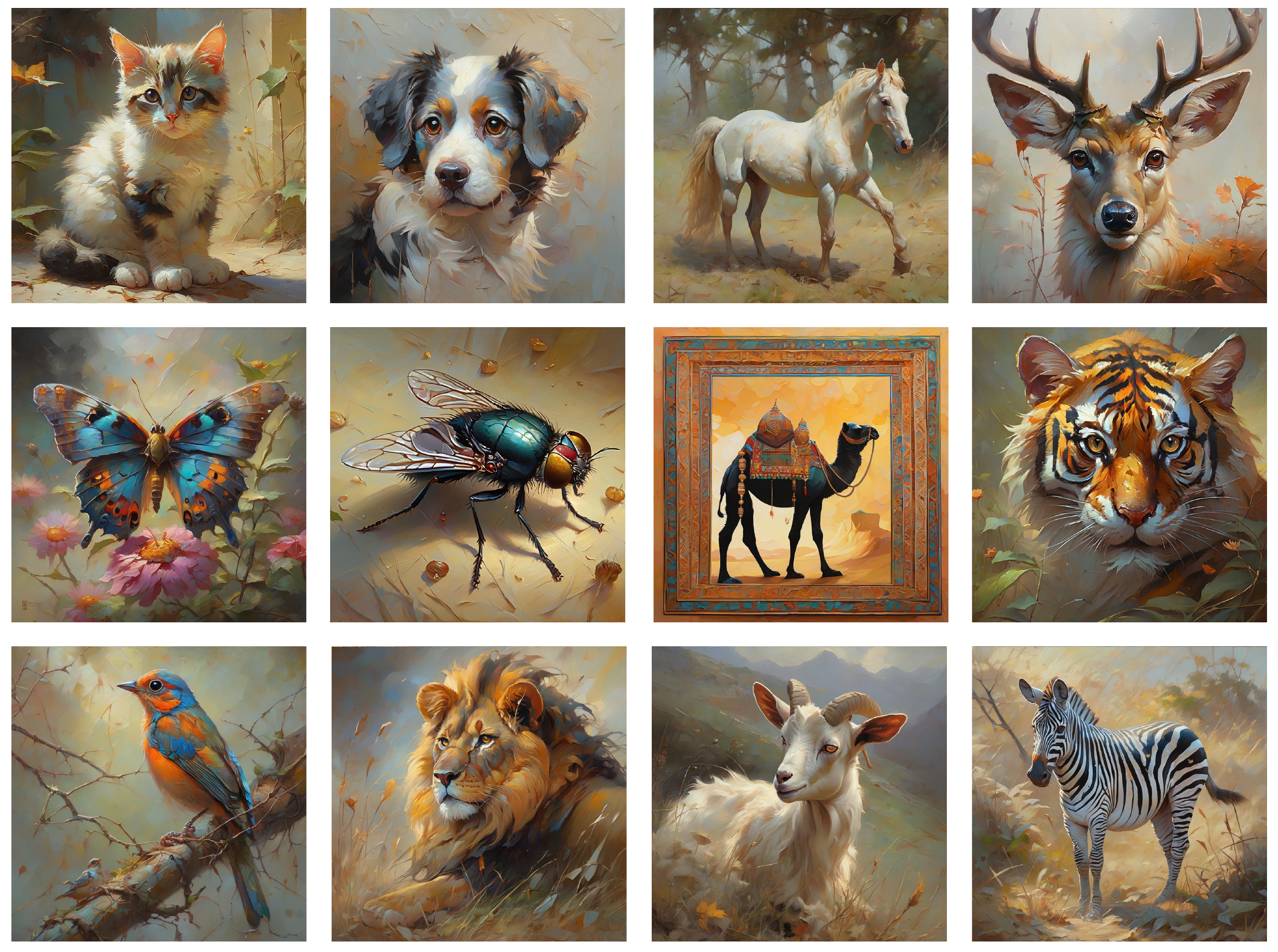}
\caption{\textbf{Uncurated samples from fine-tuning SDXL on Aesthetic Score as the reward.}}
\label{figure:appendix_SDXL_aes}
\end{figure}

\begin{figure}[t]
\centering
\includegraphics[scale=0.33]{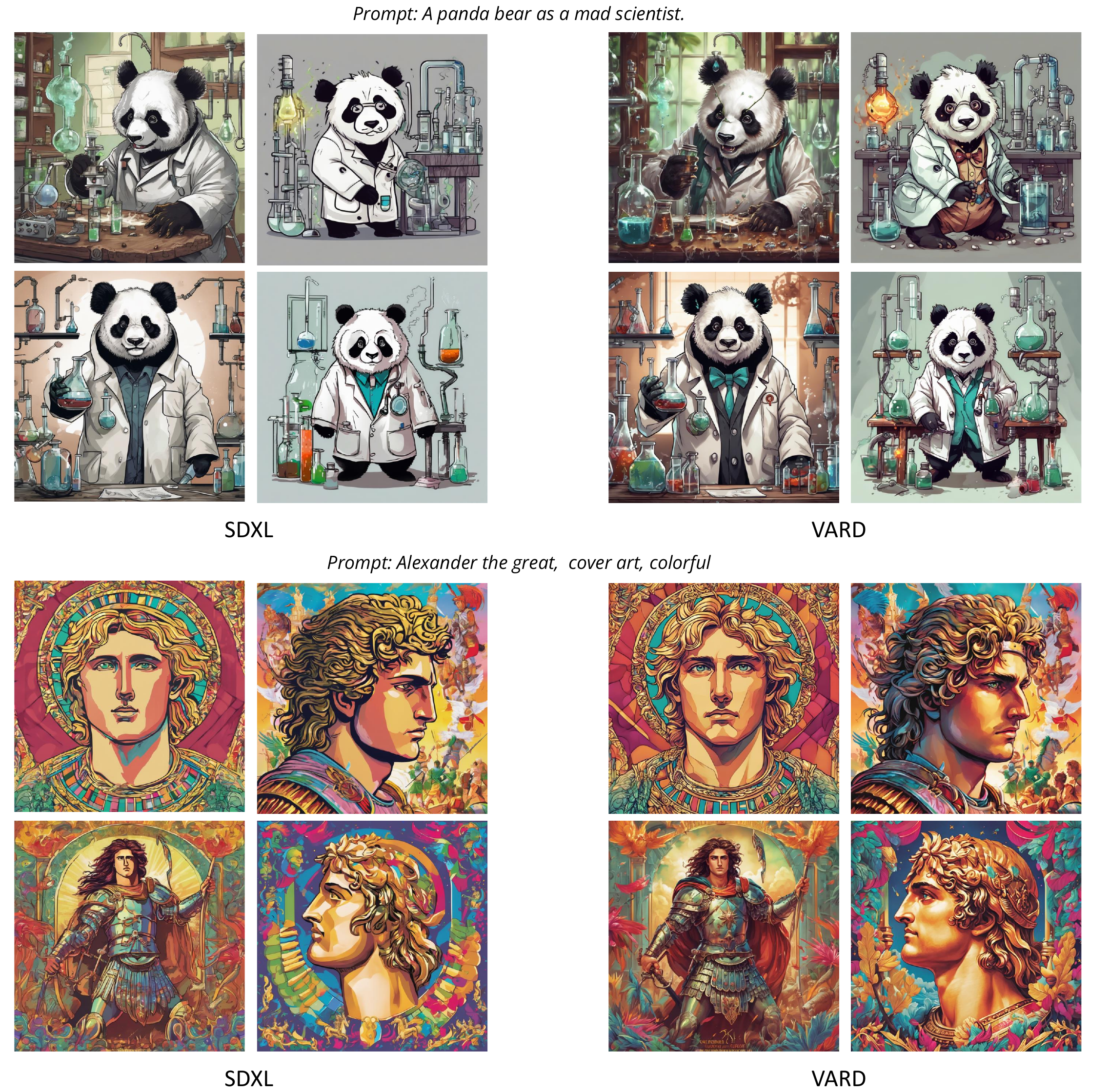}
\caption{\textbf{Further comparison of the SDXL and VARD fine-tuned version with PickScore as reward.}}
\label{figure:appendix_sdxl_pickscore}
\end{figure}

\begin{figure}[t]
\centering
\includegraphics[scale=0.33]{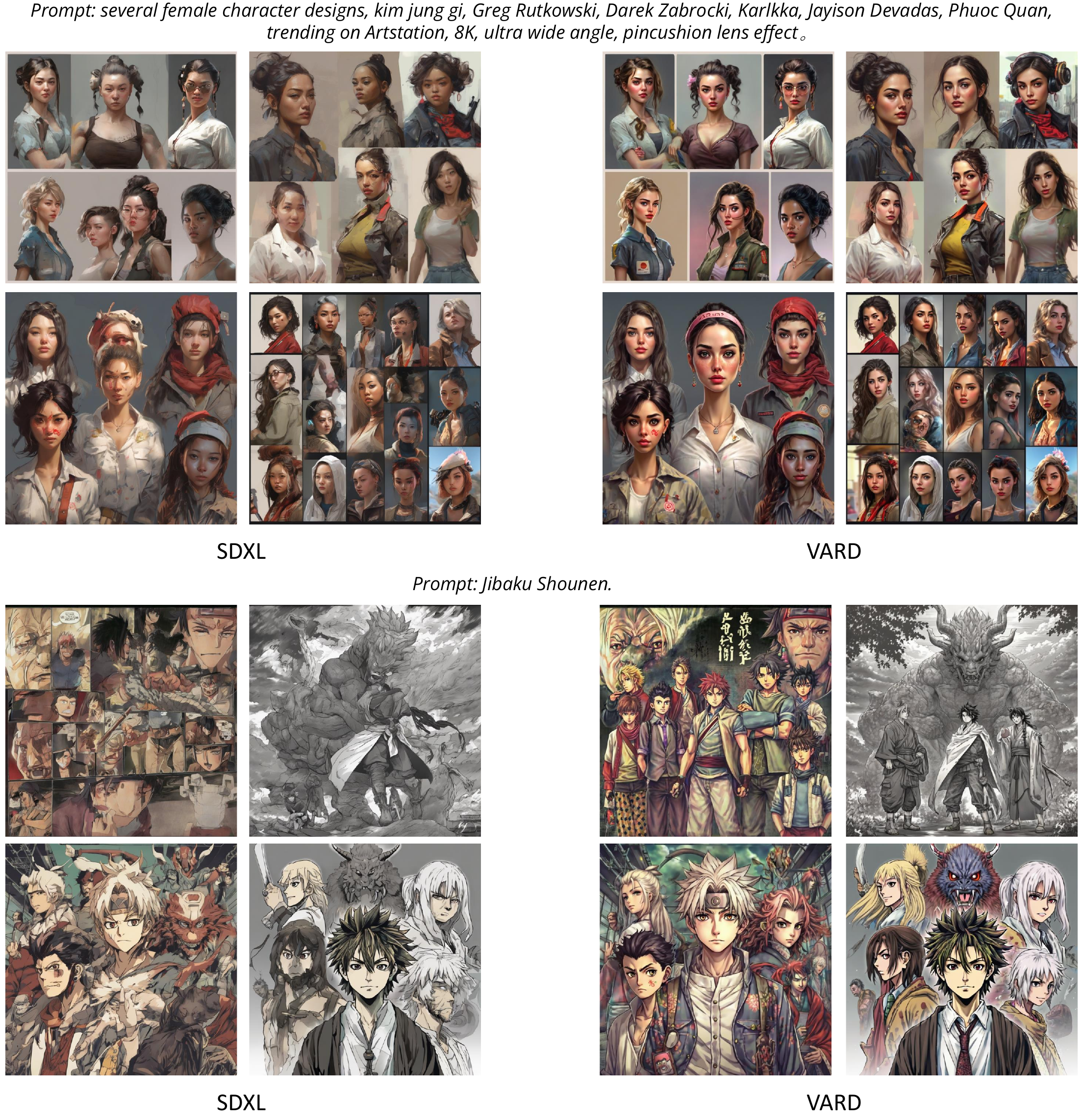}
\caption{\textbf{Further comparison of the SDXL and VARD fine-tuned version with ImageReward as reward.}}
\label{figure:appendix_sdxl_imagereward}
\end{figure}

\begin{figure}[t]
\centering
\includegraphics[scale=0.33]{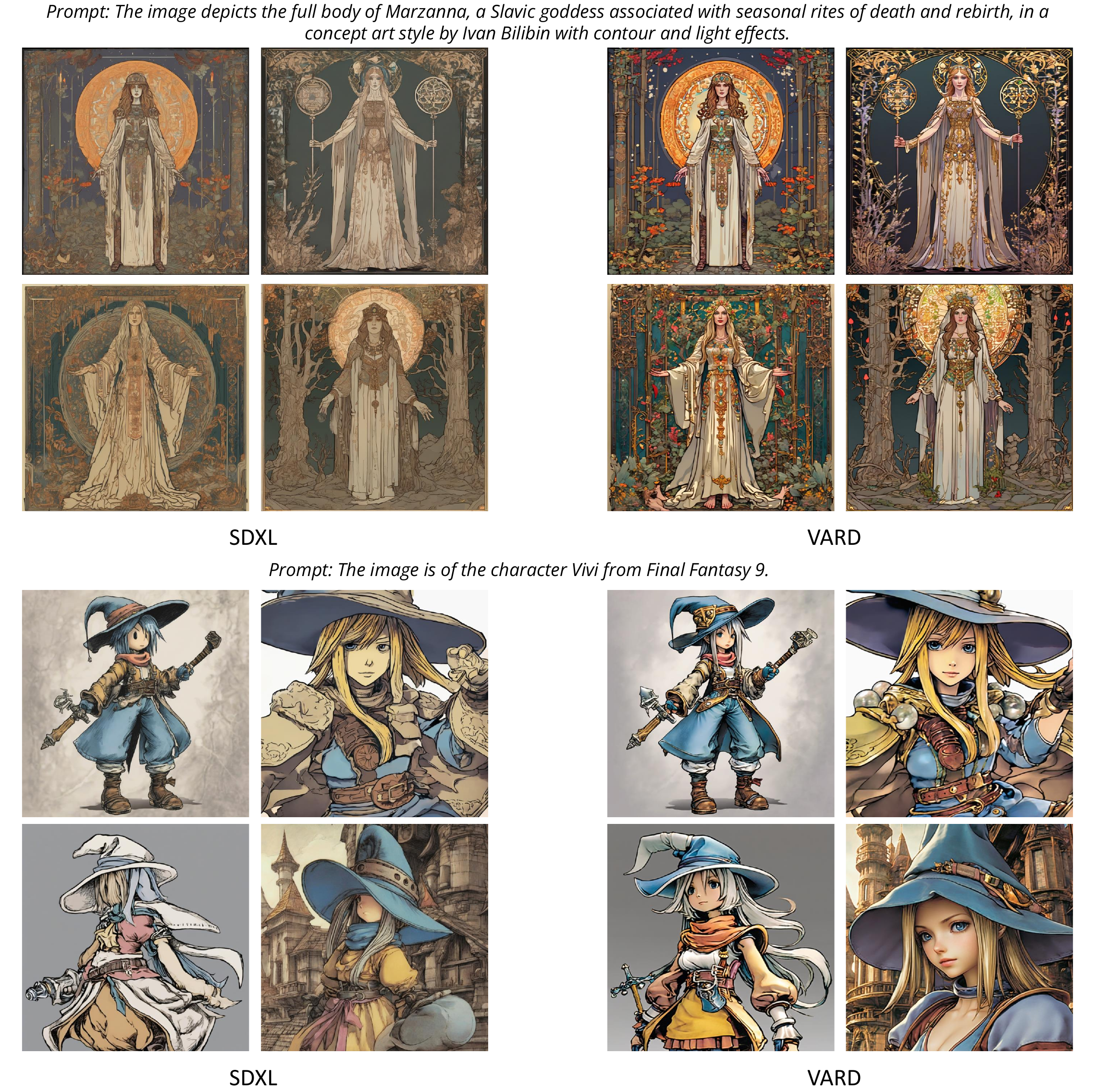}
\caption{\textbf{Further comparison of the SDXL and VARD fine-tuned version with HPSv2 as reward.}}
\label{figure:appendix_sdxl_hpsv2}
\end{figure}

We conducted these experiments using fp16 precision for training efficiency, reverting to float32 solely for the VAE decoding stage to ensure numerical stability. Given that SDXL has a significantly larger parameter count compared to SD1.4, we employed Low-Rank Adaptation (LoRA) for the fine-tuning process. Uncurated samples are presented in
Figure~\ref{figure:appendix_SDXL_aes}, Figure~\ref{figure:appendix_sdxl_pickscore}, Figure~\ref{figure:appendix_sdxl_imagereward} and Figure~\ref{figure:appendix_sdxl_hpsv2}.

\begin{figure}[t]
\centering
\includegraphics[scale=0.33]{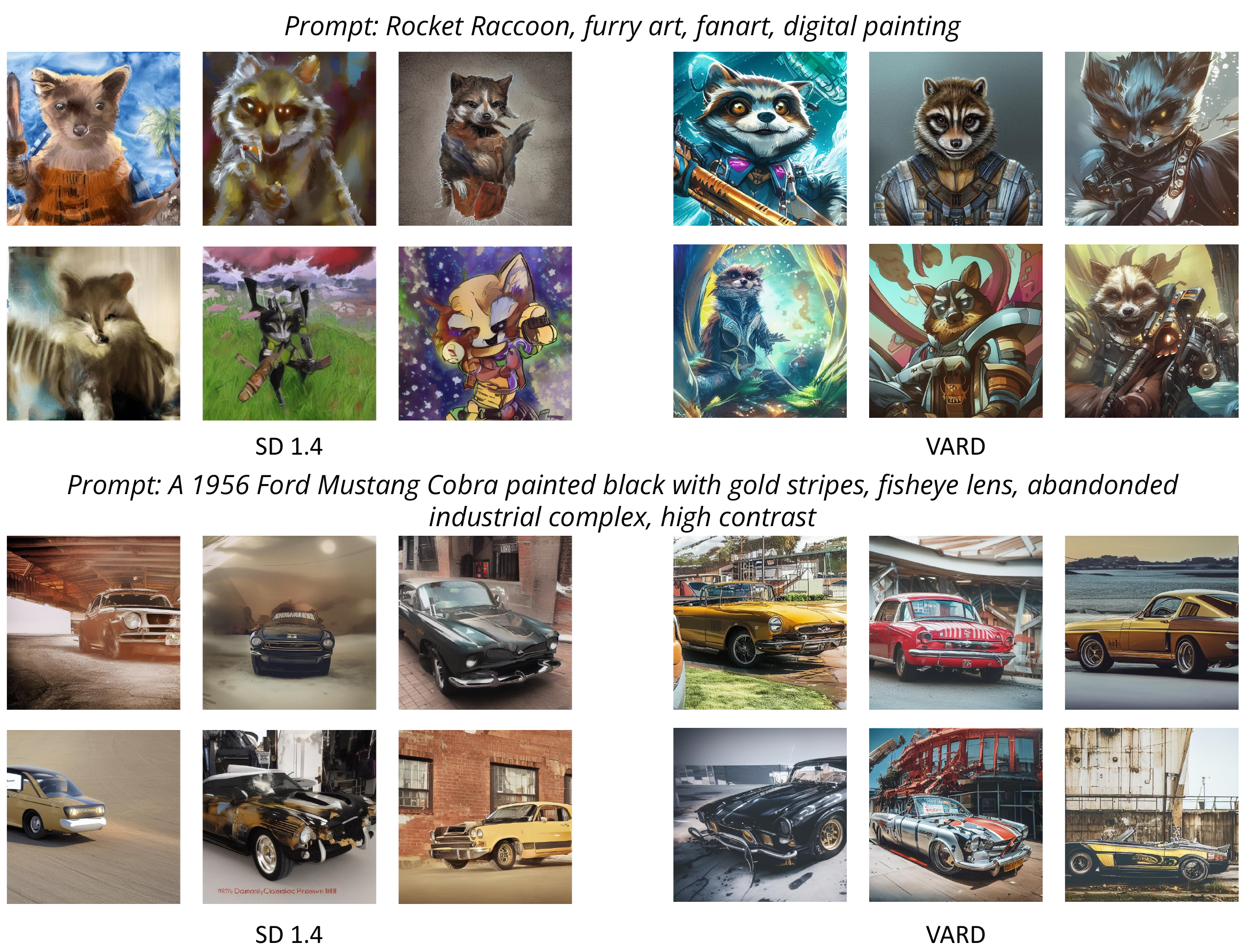}
\caption{\textbf{Further comparison of the base model and VARD fine-tuned version with PickScore as reward.}}
\label{figure:appendix_pickscore}
\end{figure}

\end{document}